\documentclass[journal]{IEEEtran}

\usepackage[utf8]{inputenc}
\usepackage[english]{babel}
\usepackage{lineno}
\usepackage{amssymb, amsmath, mathtools,bbm}
\usepackage{amsfonts,amssymb,amsthm,epsfig,url,epstopdf,array}
\usepackage{graphicx}
\usepackage[justification=centering]{caption}
\usepackage{multirow}
\usepackage{hyperref}
\usepackage{enumerate}
\usepackage{polski}
\usepackage{color}
\usepackage[pdftex,dvipsnames]{xcolor}
\usepackage{float}
\usepackage{subcaption}
\usepackage{booktabs,caption}
\usepackage{xargs}
\usepackage{pgf}
\usepackage{tikz}
\usetikzlibrary{arrows,automata,positioning,shapes.geometric}
\usepackage{relsize}
\usepackage{algorithm}
\usepackage{algpseudocode}
\usepackage{caption}
\usepackage{tablefootnote}
\usepackage{threeparttable}
\usepackage{cite}

\usepackage{booktabs,caption}
\usepackage{ifpdf}
\usepackage{algorithm}
\usepackage{algpseudocode}
\usepackage{array}

\algrenewcommand\algorithmicindent{1.0em}

\begin{document}

\title{Recurrence-Aware Long-Term Cognitive Network for Explainable Pattern Classification}

\author{Gonzalo~N\'apoles,~Yamisleydi~Salgueiro,~Isel~Grau~and~Maikel~Leon~Espinosa
\thanks{Gonzalo N\'apoles is with the Department of Cognitive Science \& Artificial Intelligence, Tilburg University, The Netherlands. e-mail: g.r.napoles@uvt.nl}
\thanks{Yamisleydi Salgueiro is with the Department of Computer Science, Faculty of Engineering, Universidad de Talca, Campus Curic\'o, Chile.}
\thanks{Isel Grau is with the Information Systems Group, Department of Industrial Engineering and Innovation Sciences, Eindhoven University of Technology, The Netherlands. Part of this work was done while at the Artificial Intelligence Lab, Vrije Universiteit Brussel, Belgium.}
\thanks{Maikel Leon Espinosa is with the Department of Business Technology, Miami Herbert Business School, University of Miami, USA.}}

\maketitle

\begin{abstract}
Machine learning solutions for pattern classification problems are nowadays widely deployed in society and industry. However, the lack of transparency and accountability of most accurate models often hinders their safe use. Thus, there is a clear need for developing explainable artificial intelligence mechanisms. There exist model-agnostic methods that summarize feature contributions, but their interpretability is limited to predictions made by black-box models. An open challenge is to develop models that have intrinsic interpretability and produce their own explanations, even for classes of models that are traditionally considered black boxes like (recurrent) neural networks. In this paper, we propose a Long-Term Cognitive Network for interpretable pattern classification of structured data. Our method brings its own mechanism for providing explanations by quantifying the relevance of each feature in the decision process. For supporting the interpretability without affecting the performance, the model incorporates more flexibility through a quasi-nonlinear reasoning rule that allows controlling nonlinearity. Besides, we propose a recurrence-aware decision model that evades the issues posed by the unique fixed point while introducing a deterministic learning algorithm to compute the tunable parameters. The simulations show that our interpretable model obtains competitive results when compared to state-of-the-art white and black-box models.
\end{abstract}


\section{Introduction}
\label{sec:introduction}

Pattern recognition techniques aim to find regularities in data stored in databases or produced by signals, processes, etc. \cite{Duda2012a}. Due to the abundance of data and the increase in computational power, machine learning algorithms have a prominent role in pattern recognition applications. Overall, pattern classification focuses on assigning a label or category to each data point. Ground truth information is necessary for learning such a mapping from input data to labels. Several machine learning algorithms have proven successful in creating classification models with high accuracy, such as support vector machines, random forests, ensembles or (deep) neural networks. 

However, the ubiquitousness of machine learning algorithms deployed in today's society has raised concerns about their accountability and transparency \cite{gunning2019darpa,goodman2017european}. For most high-stakes decision problems having an accurate model is not sufficient; some degree of interpretability is also needed. As stated in \cite{Shin2020}, when users perceive that an algorithm is fairer, more accountable, transparent, and explainable, they see it as a more trustworthy and useful resource. The form of this interpretability, either as a global holistic view of the model or local explanations over particular predictions, depends on the audience of the model and the domain \cite{BARREDOARRIETA202082}. 

Global interpretability can be obtained by using intrinsically interpretable machine learning techniques, which rely on their levels of transparency \cite{Lipton2016}. Linear or logistic regression models are the simplest interpretable predictors producing explanations about the role of the features \cite{molnar2019}. Decision trees and decision lists are generally accurate predictors that can provide intrinsic interpretability when the structure is kept on a simulatable size \cite{grau2020}. On the other hand, model-agnostic post-hoc explanation methods compute local explanations from the black-box predictions to preserve their accuracy. For example, the SHAP approximation \cite{NIPS2017_7062} of Shapley values explains the role of the features in the prediction of a given instance. Another example is the local surrogate model LIME \cite{ribeiro2016}, which describes the vicinity of the prediction with a linear regression, leveraging its intrinsic interpretability but limited to a particular region of the domain. 

Overall, model-agnostic post-hoc methods generate explanations that are local or limited to feature attribution. Explanations provided by intrinsically interpretable models are derived from their structure and easily mappable to the problem domain. The author in \cite{Rudin2019} accentuates the difference between explaining the predictions of a black box and the inherent explanations of the transparent models. Moreover, they argue that the community should focus on the latter to avoid unreliable explanations and potentiate explanations that are faithful to what the model actually computes. In \cite{Shin2021b}, the effects of anthropomorphic explanations are studied, and how certain recommendations provided by systems afford humanness, which then influence trust and emotional assurance. Although we consider that local explanations can be useful for some domains, we see the development of accurate models with inherent interpretability as an open challenge.

A type of recurrent neural network with high potential for intrinsic interpretability is Fuzzy Cognitive Maps (FCM) \cite{Kosko1986}. FCMs allow modeling complex systems in terms of causal relationships and well-defined concepts. In these networks, the experts should provide the concepts defining the system and the weights connecting such concepts, although the weights could also be computed from data using learning algorithms. In general, neural networks operate like black boxes, where hidden neurons and connections do not involve any clear meaning for the problem itself. In contrast, neural concepts in FCM-based models and their connections have a precise meaning for the system under analysis and can help explain why a solution is suitable for a given problem. FCMs have been extensively applied to modeling complex systems from engineering, environmental sciences, behavioral sciences, medicine, business, and other domains \cite{PapageorgiouS13}. 

However, while FCM-based models have proven effective in scenario simulation and time series forecasting, their performance on pattern classification problems is arguable (Section \ref{sec:literature} will revise prominent models reported in the literature). There are several reasons explaining the moderate performance of FCM-based classifiers. Firstly, the network topology depends on the problem domain since hidden neurons are not allowed. Secondly, if the network converges to a unique fixed-point attractor, then the model will be able to recognize only one decision class (not necessarily the majority one). The authors in \cite{Napoles2014} prove that, under some properties of the weight matrix, an FCM with no input neurons converges to a single attractor regardless the initial concept values. As an example, the authors in \cite{Napoles2014} show several FCMs modeling the classification of drug resistance in HIV protein sequences that converge to a unique fixed point, therefore only predicting the ``susceptible'' decision class, which is the minority one. Thirdly, both the neurons' activation values and the (causal) weights are confined to a closed interval, thus limiting the coverage of the activation space \cite{Concepcion2020a}. Finally, there is a lack of learning algorithms with a strong mathematical foundation.

To tackle the last two issues, N\'apoles et al. \cite{Napoles2020a} proposed the Long-term Cognitive Networks (LTCNs). In this FCM-like model, the weights are not constrained to any specific interval, and the tunable parameters are computed using a nonsynaptic backpropagation algorithm. However, LTCNs will not necessarily produce good prediction rates in pattern classification problems either. On the one hand, the nonsynaptic learning method assumes that the domain expert is able to define the weight matrix. On the other hand, the network's convergence to a unique fixed point continues to be a serious problem.

In this paper, we propose an LTCN-based model for interpretable pattern classification in structured data, i.e., tabular datasets with well-defined features. This model solves the remaining LTCNs' issues while preserving the network's interpretability as much as possible. Overall, our proposal brings four main theoretical contributions. Firstly, we introduce a parametric quasi-nonlinear reasoning rule that allows controlling nonlinearity. Secondly, we present a recurrence-aware decision model for multi-class pattern classification. This decision model is not affected by the unique fixed point when the network converges. Thirdly, we propose a two-step learning procedure to adjust the weights in a deterministic way. The first step is unsupervised and computes the weights connecting the inner neurons (the ones mapping the problem features). The second step is supervised and computes the weights connecting the inner neurons with the decision ones in the recurrence-aware model. Finally, we describe a measure to quantify the relevance of each feature in the decision process as a mechanism to provide explanations that are directly extracted from the model.

The remainder of the paper is organized as follows. Section \ref{sec:ltcn} presents foundations of the LTCN model, starting from the classic FCM formalism. Section \ref{sec:literature} revises a selection of prominent pieces of research devoted to FCM-based classifiers. Section \ref{sec:proposal} encloses the contributions of this paper, which include the quasi-nonlinear reasoning, the recurrence-aware architecture, its learning algorithm and the feature relevance measure. Section \ref{sec:simulations} conducts extensive numerical simulations, whereas \ref{sec:conclusions} presents some concluding remarks.

\section{Long-term Cognitive Networks}
\label{sec:ltcn}

The LTCN model has its roots in the FCMs, which were originally introduced in \cite{Kosko1986} as a knowledge-based methodology for modeling complex systems. From a connectionist viewpoint, FCMs can be seen as recurrent neural networks consisting of neural concepts and signed weighted connections. Neural concepts represent variables, states, entities related to the physical system under investigation. The signed weight associated with each connection denotes the strength of the causality between the corresponding neurons. Causal relations are quantified in the $[-1, 1]$ interval, while  neurons' activation values can take values in either $[0,1]$ or $[-1, 1]$ depending on the nonlinear transfer function attached to each neuron.

In each iteration, an FCM model produces an activation vector $A_k^{(t)}=[a_{k1}^{(t)}, \ldots, a_{ki}^{(t)}, \ldots, a_{kM}^{(t)}]$ where $a_{ki}^{(t)}$ is the activation value of the $i$-th neural entity in the $t$-th iteration, given the $k$-th initial stimulus. Equation \eqref{eq:rule1} displays the recurrent reasoning rule of this model,

\begin{equation}
\label{eq:rule1}
A_k^{(t)} = f \left(A_k^{(t-1)}W \right)
\end{equation}

\noindent where $M$ denotes the number of neurons and $W_{M \times M}$ is the weight matrix such that $w_{ji}$ represents the weight connecting the $C_j$ and $C_i$ neurons, while $f(\cdot)$ is the transfer function used to keep the neurons' activation values within the allowed activation interval.

Equation \eqref{eq:rule2} presents another reasoning rule that takes into account both the neuron's previous activation value and the states of connected neurons,

\begin{equation}
\label{eq:rule2}
A_k^{(t)} = f \left( A_k^{(t-1)}W + A_k^{(t-1)}\right).
\end{equation}

The neurons' activation values are iteratively updated until (i) the network converges to a fixed-point attractor or (ii) a maximal number of iterations $T$ is reached. These states can be defined as follows:

\begin{itemize}
  \item \textbf{Fixed point} $(\exists t_{\alpha} \in \{1,2,\dots,(T-1)\} : A_k^{(t+1)}=A_k^{(t)}, \forall k, \forall t \geq t_{\alpha})$: the network produces the same state after $t_{\alpha}$, so $A_k^{(t_{\alpha})}=A_k^{(t_{\alpha}+1)}=A_k^{(t_{\alpha}+2)}=\dots=A_k^{(T)}$. The fixed point may be unique, which means that the network will produce the same state regardless of the neuron's initial values.
  \item \textbf{Limit cycle} $(\exists t_{\alpha},P \in \{1,2,\dots,(T-1)\} : A_k^{(t+P)}=A_k^{(t)}, \forall k, \forall t \geq t_{\alpha})$: the network produces the same state periodically after the period $P$, so $A_k^{(t_{\alpha})}=A_k^{(t_{\alpha}+P)}=A_k^{(t_{\alpha}+2P)}=\dots=A_k^{(t_{\alpha}+jP)}$ where $t_{\alpha}+jP \leq T$, such that $j \in \{1,2,\dots,(T-1)\}$.
  \item \textbf{Chaos}: the network continues to produce different states for successive iterations.
\end{itemize}

These neural networks have proven effective for modeling complex systems, but they have been linked to serious misconceptions and theoretical issues \cite{NapolesSFFEVBV19}. Firstly, the fuzzy aspect in these models is ill-defined. The reasoning process of FCMs involves no fuzzy operations whatsoever, as one would expect. Secondly, FCM-based models derived from historical data (using supervised learning algorithms) can hardly be considered causal. Instead, weights resulting from data-driven construction models should be interpreted as coefficients in a regression model. Last but not least, the constraint that $w_{ji} \in [-1,1]$ greatly hinders the predictive power of FCM-based models \cite{Concepcion2020a}. Perhaps that is the reason for their scarce popularity when compared to other recurrent neural networks.

The LTCN model was introduced in \cite{Napoles2020a} to overcome these issues. In short, LTCNs are neither causal nor fuzzy, and their weights can take values in the real domain. The main similarities between FCMs and LTCNs are that they do not allow for hidden neurons (to retain the model's interpretability) and share the same recurrent reasoning rule. These features make these methods somewhat similar at first sight, but the semantic differences make them quite distinct in both theory and practice.



\section{Literature review}
\label{sec:literature}

The literature reports several works related to the use of FCM-based models in pattern classification problems. One of the first attempts at incorporating FCMs in pattern classification applications is found in \cite{PAPAKOSTAS2008}. In that paper, three classification models were proposed, but not all of them achieved good results.


The mentioned study continued with \cite{Papakostas2010} where the performances of the FCM classifiers were studied more in-depth. The authors investigated FCM-based classifiers' performance when adjusting the appropriate set of parameters, such as the transfer function, the reasoning rule, and the network topology. The newly introduced classifiers presented better prediction capabilities. However, these sophisticated models are no longer interpretable since they were hybridized with black boxes. 

The study in \cite{Song2011} presented an approach that translates the reasoning mechanism of traditional FCMs to a set of fuzzy IF–THEN rules. Each fuzzy rule is defined as a fuzzy set concerning the summation of weighted membership grades of input linguistic terms. The impacts of fuzzy rules on output linguistic parts are transferred along with fuzzy weights and quantified by mutual subsethood. The consequent parts are then defuzzified by standard-volume-based centroid defuzzification. Finally, by describing each output as a linear combination of the defuzzified consequent parts, the model takes advantage of the mapping capability offered by the consequent parts to approximate the desired outputs. Recent studies, as found on \cite{AnagnostisTAPKM21}, continued on the idea of creating models based on neuro-fuzzy inference systems with the advantage of obtaining the weights of connecting links to adjust the parameters of fuzzy rules. In other words, for determining the rules and obtaining the weights, in addition to the knowledge of experts, the model also exploits the existing data to adjust the inference system's parameters.

Separately, the need to apply learning algorithms for training FCMs was discussed in \cite{PapageorgiouSG04} and \cite{Papakostas2012}. Hebbian-like algorithms aimed to adjust the weights between the neurons of the FCM classifier so that it can converge to the desired state. The works in \cite{PapageorgiouG05}, \cite{StachKP10}, \cite{StachPK12}, \cite{Len2013} and \cite{SalmeronRM17} reported the use of different paradigms deviating from Hebbian-based learners and employing other approaches (e.g., based on evolutionary algorithms).

In a different direction, the work in \cite{Npoles2016} extended the use of FCMs with the creation of Rough Cognitive Networks (RCNs) as granular classifiers stemming from the hybridization of FCMs and Rough Set Theory. Such cognitive neural networks attempted to quantify the impact of rough granular constructs over each decision class for a problem at hand. RCNs have shown substantial improvements in solving different classification problems, but the model reported some sensitivity to the similarity threshold upon which the rough information granules are built. Moreover, Fuzzy-Rough Cognitive Networks (FRCNs) \cite{Npoles2018,Concepcion2020} improved on this limitation and obtained results as accurate as of the most successful black-box models with the main advantage of being able to elucidate its decision process using inclusion degrees and causal relations.

\begin{figure}[!ht]
\centering
\captionsetup{justification=justified}
\includegraphics[width=0.49\textwidth]{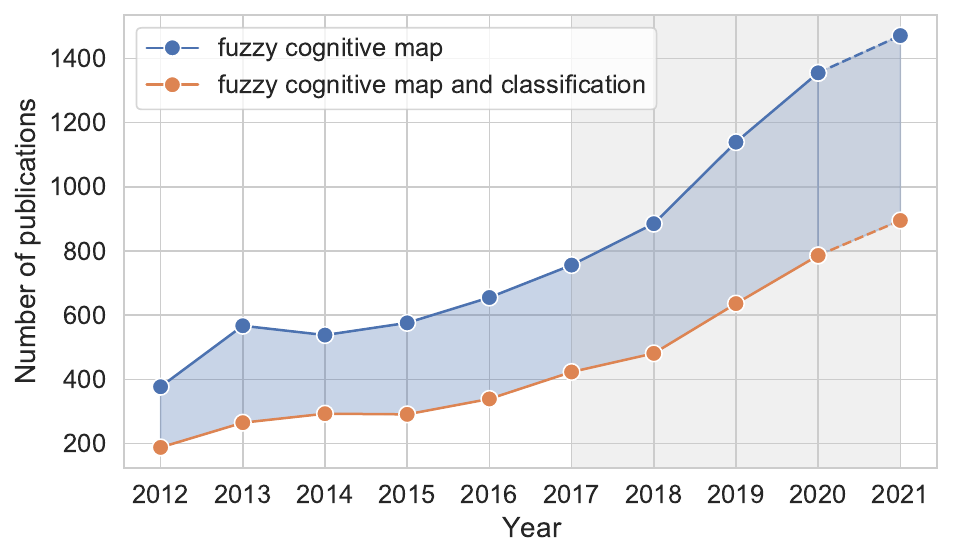}
\caption{Citation count of articles referencing to ``fuzzy cognitive maps'' and ``fuzzy cognitive maps'' + ``classification'' in papers published in the period 2012-2020, extracted from \url{https://app.dimensions.ai}. For 2021, the statistics are included in dashed lines for the sake of completeness. The period highlighted in gray signals a change in the focus of the algorithms, where more mathematically founded proposals rather than meta-heuristic inspired solutions start to appear.}
\label{fig:fcm-pub}
\end{figure}

There is little doubt that FCMs have been quite useful for designing knowledge-based systems involving experts, with their intrinsic interpretability being a pivotal feature. Overall, the development of FCM-based models and classifiers is on the rise (as seen in Figure \ref{fig:fcm-pub}). If only FCM-based models were to be as effective as black-box models, then we would have a reasoning model able to provide explanations without using any post-hoc method.

Wrapping up, most of the seen algorithms created for classification purposes dive two-fold as follows: (1) low-level (where neurons correspond to system variables) and (2) high-level (where neurons correspond to information granules). The model to be presented in this paper belongs to the first class. Many of the limitations observed correspond to open problems \cite{Npoles2017a}, e.g., the convergence to a unique fixed point might mean the recognition of only one class. Also, a considerable number of learning algorithms for FCMs reported in the literature are meta-heuristic-based, therefore likely to suffer from both speed and convergence problems, e.g., easily converge to local optima. While the literature review shows substantial progress on FCMs and their use for solving pattern classification problems, the above-mentioned drawbacks serve as a motivation to develop our proposal.

\section{Recurrence-aware LTCN-based classifier}
\label{sec:proposal}

This section presents a recurrence-aware LTCN-based model that allows for explainable pattern classification. The first subsection introduces a quasi-nonlinear reasoning rule that uses a parameter to control the nonlinearity degree of the recurrent reasoning process. In the second subsection, we explain our model's architecture and detail a two-step learning procedure to estimate the tunable parameters in a deterministic way. Finally, we introduce a measure to quantify the relevance of each problem feature in the classifier's decision process.

\subsection{Quasi-nonlinear reasoning model}
\label{sec:proposal:rule}

Firstly, we propose a new reasoning rule that introduces a nonlinearity coefficient $\phi \in [0,1]$ controlling the extent to which the model will take into account the value produced by the transfer function over the neuron's initial activation value. Equation \eqref{eq:rule3} shows this model,

\begin{equation}
\label{eq:rule3}
A_k^{(t)} = \phi f \left( A_k^{(t-1)}W + B \right)+ (1-\phi) A_k^{(0)}
\end{equation}

\noindent where $B_{1 \times M}$ denotes the bias matrix, which can be understood as the amount of external information impacting the neuron's state (i.e., what cannot be explained through the neuronal concepts describing the problem domain). The matrices $B$ and $W$ will be computed from historical data during the \textit{unsupervised learning step}.

If $\phi=1$, then we have a traditional long-term recurrent model such that the neurons' activation values depend on the states of connected neurons in the previous iteration. If $\phi=0$, then there will be no recurrence at all, so the model will narrow down to a linear regression with multiple outputs \cite{Yu2019}. For the most part, the motivation for this model is that in traditional FCMs, the input is explicitly used to compute neurons' activation values only in the first iteration. Moreover, some decision models might benefit from a certain degree of linearity.

\begin{figure}[!htbp]
\center

    \begin{subfigure}{0.49\textwidth}
	\center
	\resizebox{\columnwidth}{!}{
	\begin{tikzpicture}[->,>=stealth',shorten >=1pt,auto,node distance=2cm,main node/.style={minimum size=8mm,circle,draw=RoyalBlue,font=\sffamily\bfseries}]
	
	\node[main node] (1) [font=\small, thick, draw=RoyalBlue] {$\bar{a}_{ki}^{(t)}$};
	
	\node[main node] (2) [font=\small, thick, draw=RoyalBlue, above left = 2cm of 1] {$a_{k1}^{(t-1)}$};
	\node[main node] (3) [font=\small, thick, draw=RoyalBlue, left = 1cm of 1] {$a_{k2}^{(t-1)}$};
	\node[main node] (4) [font=\small, thick, draw=none, below = 0.1325cm of 3] {$...$};
	\node[main node] (5) [font=\small, thick, draw=RoyalBlue, below = 1.06cm of 3] {$a_{kM}^{(t-1)}$};
	
	\node[main node] (10) [font=\small, thick, diamond, draw=RoyalBlue, right = 1cm of 1] {$f(\cdot)$};
	
	\node[main node] (11) [font=\small, thick, draw=RoyalBlue, right = 1cm of 10] {$a_{ki}^{(t)}$};
	
	\node (in1) [font=\small, thick, draw=none, left = 0.5cm of 2] {};
	\node (in2) [font=\small, thick, draw=none, left = 0.5cm of 3] {};
	\node (inm) [font=\small, thick, draw=none, left = 0.5cm of 5] {};
	\node (out) [font=\small, thick, draw=none, right = 0.5cm of 11] {};
	
	\path[every node/.style={ font=\sffamily}]
	(1) edge node [above] {} (10)
	
	(2) edge node [above, sloped] {$w_{1i}$} (1)
	(3) edge node [above, sloped] {$w_{2i}$} (1)
	(5) edge node [above, sloped] {$w_{Mi}$} (1)
	
	(in1) edge node [above, sloped] {} (2)
	(in2) edge node [above, sloped] {} (3)
	(inm) edge node [above, sloped] {} (5)
	(11) edge node [above, sloped] {} (out)

    (10) edge node [above] {} (11)
	;
    \end{tikzpicture}
    }
	\caption{Reasoning rule in Equation \eqref{eq:rule1}}
	\label{fig:act_rule1}
	\end{subfigure}
	
	\vspace{3mm}
	
	\begin{subfigure}{0.49\textwidth}
	\center
	\resizebox{\columnwidth}{!}{
	\begin{tikzpicture}[->,>=stealth',shorten >=1pt,auto,node distance=2cm,main node/.style={minimum size=8mm,circle,draw=RoyalBlue,font=\sffamily\bfseries}]
	
	\node[main node] (1) [font=\small, thick, draw=RoyalBlue] {$\bar{a}_{ki}^{(t)}$};
	
	\node[main node] (2) [font=\small, thick, draw=RoyalBlue, above left = 2cm of 1] {$a_{k1}^{(t-1)}$};
	\node[main node] (3) [font=\small, thick, draw=RoyalBlue, left = 1cm of 1] {$a_{k2}^{(t-1)}$};
	\node[main node] (4) [font=\small, thick, draw=none, below = 0.1325cm of 3] {$...$};
	\node[main node] (5) [font=\small, thick, draw=RoyalBlue, below = 1.06cm of 3] {$a_{kM}^{(t-1)}$};
	\node[main node] (6) [font=\small, thick, draw=RoyalBlue, below = 1.2cm of 1] {$a_{ki}^{(t-1)}$};
	
	\node[main node] (10) [font=\small, thick, diamond, draw=RoyalBlue, right = 1cm of 1] {$f(\cdot)$};
	
	\node[main node] (11) [font=\small, thick, draw=RoyalBlue, right = 1cm of 10] {$a_{ki}^{(t)}$};
	
	\node (in1) [font=\small, thick, draw=none, left = 0.5cm of 2] {};
	\node (in2) [font=\small, thick, draw=none, left = 0.5cm of 3] {};
	\node (inm) [font=\small, thick, draw=none, left = 0.5cm of 5] {};
	\node (out) [font=\small, thick, draw=none, right = 0.5cm of 11] {};
	
	\path[every node/.style={ font=\sffamily}]
	(1) edge node [above] {} (10)
	
	(2) edge node [above, sloped] {$w_{1i}$} (1)
	(3) edge node [above, sloped] {$w_{2i}$} (1)
	(5) edge node [above, sloped] {$w_{Mi}$} (1)
	(6) edge node [below, sloped] {} (10)
	
	(in1) edge node [above, sloped] {} (2)
	(in2) edge node [above, sloped] {} (3)
	(inm) edge node [above, sloped] {} (5)
	(11) edge node [above, sloped] {} (out)

    (10) edge node [above] {} (11)
	
	;
    \end{tikzpicture}
    }
	\caption{Reasoning rule in Equation \eqref{eq:rule2}}
	\label{fig:act_rule2}
	\end{subfigure}
	
	\vspace{3mm}
	
	\begin{subfigure}{0.49\textwidth}
	\centering
	\resizebox{\columnwidth}{!}{
	\begin{tikzpicture}[->,>=stealth',shorten >=1pt,auto,node distance=2cm,main node/.style={minimum size=8mm,circle,draw=RoyalBlue,font=\sffamily\bfseries}]
	
	\node[main node] (1) [font=\small, thick, draw=RoyalBlue] {$\bar{a}_{ki}^{(t)}$};
	
	\node[main node] (2) [font=\small, thick, draw=RoyalBlue, above left = 2cm of 1] {$a_{k1}^{(t-1)}$};
	\node[main node] (3) [font=\small, thick, draw=RoyalBlue, left = 1.03cm of 1] {$a_{k2}^{(t-1)}$};
	\node[main node] (4) [font=\small, thick, draw=none, below = 0.1325cm of 3] {$...$};
	\node[main node] (5) [font=\small, thick, draw=RoyalBlue, below = 1.06cm of 3] {$a_{kM}^{(t-1)}$};
	\node[main node] (6) [font=\small, thick, draw=RoyalBlue, below = 1.125cm of 10] {$a_{ki}^{(0)}$};
	
	\node[main node] (10) [font=\small, thick, diamond, draw=RoyalBlue, right = 1cm of 1] {$f(\cdot)$};
	
	\node[main node] (11) [font=\small, thick, draw=RoyalBlue, right = 1cm of 10] {$a_{ki}^{(t)}$};
	
	\node[main node] (12) [font=\small, thick, draw=RoyalBlue, minimum size=1.06cm, right = 1cm of 5] {$1.0$};
	
	\node (in1) [font=\small, thick, draw=none, left = 0.5cm of 2] {};
	\node (in2) [font=\small, thick, draw=none, left = 0.5cm of 3] {};
	\node (inm) [font=\small, thick, draw=none, left = 0.5cm of 5] {};
	\node (out) [font=\small, thick, draw=none, right = 0.5cm of 11] {};
	
	\path[every node/.style={ font=\sffamily}]
	(1) edge node [above] {} (10)
	
	(2) edge node [above, sloped] {$w_{1i}$} (1)
	(3) edge node [above, sloped] {$w_{2i}$} (1)
	(5) edge node [above, sloped] {$w_{Mi}$} (1)
	(6) edge node [below, sloped] {$1-\phi$} (11)
	
	(in1) edge node [above, sloped] {} (2)
	(in2) edge node [above, sloped] {} (3)
	(inm) edge node [above, sloped] {} (5)
	(11) edge node [above, sloped] {} (out)
	
	(12) edge node [below, sloped] {$b_i^{(t)}$} (10)

    (10) edge node [above] {$\phi$} (11)
	
	;
    \end{tikzpicture}
    }
	\caption{Reasoning rule in Equation \eqref{eq:rule3}}
	\label{fig:act_rule3}
	\end{subfigure}
\captionsetup{justification=justified}
\caption{Reasoning rules for LTCN-based models. In the first model, the neuron's activation value is determined by the states of connected neurons. In the second model, the activation value additionally considers the neuron's previous state before applying the transfer function. The aggregation of these values is passed through the transfer function. In the third model, the neuron's activation value is determined by the bias, the states of connected neurons, and the neuron's previous state. However, the previous state is added after having applied the transfer function. Overall, the $\phi$ parameter controls the amount of information we take from the transfer function output, which is regarded as nonlinear information.}
\label{fig:act_rules}
\end{figure}
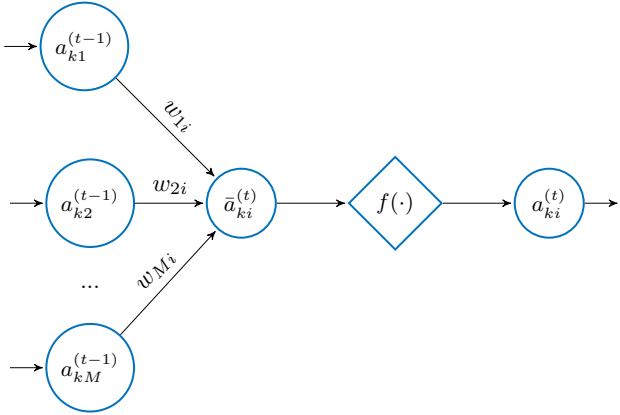
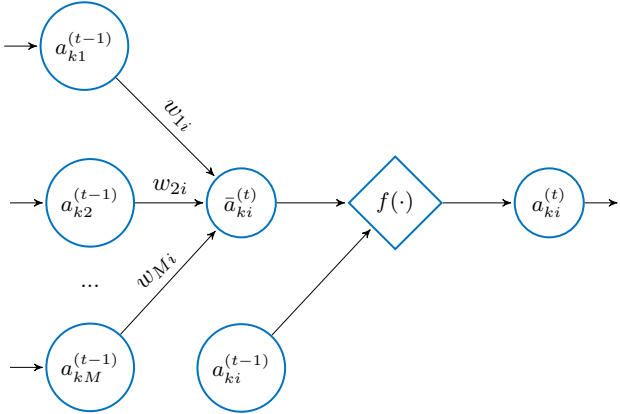
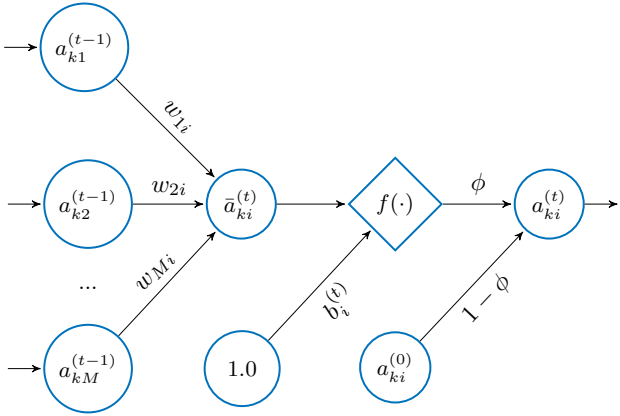

Figures \ref{fig:act_rule1}, \ref{fig:act_rule2} and \ref{fig:act_rule3} depict the reasoning rules formalized in Equations \eqref{eq:rule1}, \eqref{eq:rule2} and \eqref{eq:rule3}, respectively. Notice that, unlike the model in Figure \ref{fig:act_rule2}, our proposal adds the scaled initial activation value after having transformed the incoming information flow with the transfer function. Failing to do that implies that the additional activation value (either the neuron's initial or previous activation value) will have a limited impact on the neuron's outcome after applying the transfer function.

Overall, one can see that the proposed quasi-nonlinear reasoning rule uses the initial activation value to compute the neurons' states in each iteration. However, such a reasoning rule will not prevent the network from converging to undesirable states. An example of these states is the unique fixed-point attractor since it causes the network to produce the same outputs to any input. This issue will be discussed in the next sections.

\subsection{Network architecture and learning}
\label{sec:proposal:architecture}

Our recurrence-aware neural classifier involves two building blocks. The first one consists of an LTCN model where each neuron maps a problem variable (feature). The role of this neural block is to capture the dynamics of the system, which can be either fixed-point attractors, cyclic or chaotic states. Of course, the ideal situation for this model is for each decision class to be associated with a different equilibrium point. However, we will not make any assumptions on the convergence properties of this model. It is worth mentioning that this model will use the reasoning rule in Equation \eqref{eq:rule3}, which involves a parameter to control nonlinearity. The second building block connects the inner neurons denoting problem features with the decision neurons. The first neural block will be trained using an unsupervised learning approach, while the second learning step will be supervised.

Equation \eqref{eq:weight} displays the unsupervised learning rule to compute the $i$-th column of the weight matrix $W_{M \times M}$ and the bias $b_i$ connected to the $i$-th neuron,

\begin{equation}
\label{eq:weight}
\begin{bmatrix} b_i \\ W_i \end{bmatrix} = (L^{\top}L)^{-1} L^{\top} f^{-1}(X_i)
\end{equation}

\noindent where $X_i$ is the $i$-th column of the training set $X_{K \times M}$ and $L$ is a $K \times (M+1)$ matrix that results after replacing the $i$-th column of $X$ with zeros and concatenating a $K \times 1$ column vector full of ones, while $K$ denotes the number of training instances. Those weights correspond to the coefficients of $M$ regression models such that $X_i$ is deemed the target variable of the $i$-th model. We assume that the training set has been normalized and that they are inverse-friendly (i.e., they do not cause $f^{-1}(.)$ to produce $-\infty$, $+\infty$ or any indeterminate behavior). The intuition of this procedure is that we want to approximate the $i$-th problem variable given the remaining ones.

The second component of our proposal is a recurrence-aware sub-network that connects each temporal state $A_k^{(t)}$ with the decision neurons. This sub-network uses all states resulting from the recurrent reasoning rule for a new instance. Equation \eqref{eq:decision} shows the model used to compute the activation values of decision neurons,

\begin{equation}
\label{eq:decision}
\hat{Y}_k = f\left(H_k^{(T)} R + Q \right)
\end{equation}

\noindent where $\hat{Y}_k$ is the prediction for the $k$-th training instance, $R_{{M(T+1) \times N}}$ is the outer weight matrix connecting the temporal states (including the initial state) with the $N$ decision neurons, while $Q_{1 \times N}$ is the bias weight vector attached to decision neurons. The matrices $R$ and $Q$ will be computed from historical data during the \textit{supervised learning step}. In this formulation, $H_k^{(T)}$ is a $1 \times M(T+1)$ matrix resulting from the recursive horizontal concatenation of the $T+1$ temporal states:

\begin{equation}
\label{eq:concat}
H_k^{(t)} = \left(H_k^{(t-1)} | A_k^{(t)}\right)
\end{equation}

\noindent where $H_k^{(0)}=X_k$ while $(\cdot | \cdot)$ stands for the concatenation operator. Therefore, it holds that

\begin{equation}
\label{eq:concat}
H_k^{(T)} = \left(H_k^{(0)} | A_k^{(1)} | A_k^{(2)} | \ldots | A_k^{(T-2)} | A_k^{(T-1)} | A_k^{(T)}\right).
\end{equation}

For the sake of clarity, we have made an explicit distinction between the inner weights connecting the features, and the outer weights connecting the temporal states with the decision neurons. The same design choice applies to the inner and outer bias weights.

Figure \ref{fig:models} shows the decision model of traditional FCM-based classifiers and the one proposed in this paper. In the first case, the model narrows down to a linear regression where the final state $A_{k}^{(T)}$ is used as independent variables. However, as stated earlier, if the network converges to the unique fixed-point attractor, $A_{k}^{(T)}$ will be the same for all initial activation values. If this situation comes to light, the model will produce the same decision class, as seen in \cite{Napoles2014}. In contrast, the proposed recurrence-aware model uses all temporal states as inputs of a regression model, thus preventing the classifier from producing the same decision class when converging to the unique fixed point. Therefore, our LTCN-based model will focus on the trajectory to the fixed point instead of focusing on the equilibrium point itself. 

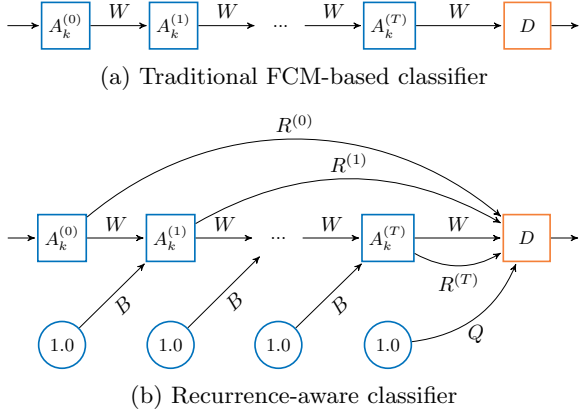
\begin{figure}[!ht]
\center

    \begin{subfigure}{0.5\textwidth}
	\center
	\resizebox{\columnwidth}{!}{
	\begin{tikzpicture}[->,>=stealth',shorten >=1pt,auto,node distance=2cm,main node/.style={minimum size=8mm,circle,draw=RoyalBlue,font=\sffamily\bfseries}]
	
	\node[main node] (0) [font=\small, thick, draw=none, left = 0.5cm of 1] {};
	
	\node[main node] (1) [font=\small, thick, rectangle, draw=RoyalBlue] {$A_{k}^{(0)}$};
	\node[main node] (2) [font=\small, thick, rectangle, draw=RoyalBlue, right = 1cm of 1] {$A_{k}^{(1)}$};
	\node[main node] (3) [font=\small, thick, draw=none, right = 1cm of 2] {$...$};
	\node[main node] (4) [font=\small, thick, rectangle, draw=RoyalBlue, right = 1cm of 3] {$A_{k}^{(T)}$};
	\node[main node] (5) [font=\small, thick, rectangle, draw=Orange, right = 1.5cm of 4] {$D$};
	
	\node[main node] (10) [font=\small, thick, draw=none, right = 0.5cm of 5] {};
	
	\path[every node/.style={ font=\sffamily}]
	(0) edge node [above] {} (1)
	(1) edge node [above] {$W$} (2)
	(2) edge node [above] {$W$} (3)
	(3) edge node [above] {$W$} (4)
	(4) edge node [above] {$W$} (5)
	(5) edge node [above] {} (10)
	
	;
    \end{tikzpicture}
    }
	\caption{Traditional FCM-based classifier}
	\label{fig:model1}
	\end{subfigure}
	
	\vspace{2mm}
	
	\begin{subfigure}{0.5\textwidth}
	\center
	\resizebox{\columnwidth}{!}{
	\begin{tikzpicture}[->,>=stealth',shorten >=1pt,auto,node distance=2cm,main node/.style={minimum size=8mm,circle,draw=RoyalBlue,font=\sffamily\bfseries}]
	
	\node[main node] (0) [font=\small, thick, draw=none, left = 0.5cm of 1] {};
	
	\node[main node] (1) [font=\small, thick, rectangle, draw=RoyalBlue] {$A_{k}^{(0)}$};
	\node[main node] (2) [font=\small, thick, rectangle, draw=RoyalBlue, right = 1cm of 1] {$A_{k}^{(1)}$};
	\node[main node] (3) [font=\small, thick, draw=none, right = 1cm of 2] {$...$};
	\node[main node] (4) [font=\small, thick, rectangle, draw=RoyalBlue, right = 1cm of 3] {$A_{k}^{(T)}$};
	\node[main node] (5) [font=\small, thick, rectangle, draw=Orange, right = 1.5cm of 4] {$D$};
	
	\node[main node] (6) [font=\small, thick, draw=RoyalBlue, below = 1cm of 1] {$1.0$};
	\node[main node] (7) [font=\small, thick, draw=RoyalBlue, below = 1cm of 2] {$1.0$};
	\node[main node] (8) [font=\small, thick, draw=RoyalBlue, below = 1cm of 3] {$1.0$};
	\node[main node] (9) [font=\small, thick, draw=RoyalBlue, below = 1cm of 4] {$1.0$};
	
	\node[main node] (10) [font=\small, thick, draw=none, right = 0.5cm of 5] {};
	
	\path[every node/.style={ font=\sffamily}]
	(0) edge node [above] {} (1)
	(1) edge node [above] {$W$} (2)
	(2) edge node [above] {$W$} (3)
	(3) edge node [above] {$W$} (4)
	(4) edge node [above] {$W$} (5)
	(5) edge node [above] {} (10)
	
	(6) edge node [below, sloped] {$B$} (2)
	(7) edge node [below, sloped] {$B$} (3)
	(8) edge node [below, sloped] {$B$} (4)
	(9) edge [bend right=30] node [below] {$Q$} (5)
	
	(1) edge [bend left=40] node [above] {$R^{(0)}$} (5)
	(2) edge [bend left] node [above] {$R^{(1)}$} (5)
	(4) edge [bend right=30] node [below] {$R^{(T)}$} (5)

	;
    \end{tikzpicture}
    }
	\caption{Recurrence-aware classifier}
	\label{fig:model2}
	\end{subfigure}

\captionsetup{justification=justified}
\caption{Decision model of a classic FCM-based classifier compared to our recurrence-aware model. In the former model, the decision class is determined from the last state. In our model, the decision class is computed considering all temporal states the network produces during the recurrent reasoning process. This makes our model less sensitive to the unique fixed-point attractor.}
\label{fig:models}
\end{figure}

The last step concerns the supervised learning approach to adjust the tunable parameters. This means that we have to estimate the outer weights (denoted with the matrix $R$) and the outer bias weights attached to decision neurons (denoted with the matrix $Q$). Equation \eqref{eq:learning} formalizes how to compute both weight matrices in a single step using the following pseudoinverse learning rule:

\begin{equation}
\label{eq:learning}
\begin{bmatrix} R \\ Q \end{bmatrix} = \left(H^{(T)} | \mathbbm{1}\right)^{\ddagger}f^{-1}(Y) 
\end{equation}

\noindent where $\mathbbm{1}$ denotes a $K \times 1$ column vector full of ones, $(\cdot)^{\ddagger}$ represents the Moore-Penrose pseudoinverse \cite{penrose_1955}, while $Y_{K \times N}$ is a matrix containing the inverse-friendly one-hot encoding of the decision classes. The Moore-Penrose pseudoinverse is computed using the orthogonal projection method. If any matrix $H$ has linearly independent columns ($H^{\top}H$ is nonsingular), then $H^\ddagger=(H^{\top}H)^{-1}H^{\top}$. In contrast, if $H$ has linearly independent rows ($HH^{\top}$ is nonsingular), then $H^\ddagger=H^{\top}(H^{\top}H)^{-1}$. The former is a left inverse because $H^\ddagger H=I$ and the latter is a right inverse because $HH^\ddagger=I$. Overall, the Moore-Penrose pseudoinverse is one of the best strategies to solve the least square problem when $H$ is not invertible.

\subsection{Feature relevance measure}
\label{sec:proposal:xai}

The advantages of our model include the ability to deal with the unique fixed point, a very low training time (to be illustrated during the experiments), and the possibility of specifying the nonlinearity degree of the reasoning model. Another reason for using FCM-based classifiers is their intrinsic interpretability. However, this does not mean we should expect to understand the model as a whole in the same way we could not easily visualize a large decision tree. Instead, we should focus on explaining the classifier's decision process by using its knowledge structures (i.e., inner and outer weights) computed during the unsupervised and supervised learning steps.

In this subsection, we will introduce a comprehensive measure to estimate the relevance of problem features for the classifier's decision process. It is worth recalling that our neural system does not include hidden neurons that might hinder its interpretability. Instead, it has meaningful neural concepts having temporal states. In other words, each component in the network has a well-defined meaning for the problem domain being modeled. Such a characteristics is pivotal for designing an intrinsic feature relevance score in our recurrence-aware classifier.

Before presenting our relevance score, we should clarify that neural concepts are not the same as problem features. While the latter are often static entities, the former change their states as the FCM model iterates. However, neural concepts can be used as proxies to quantify the relevance of features in the network's decision process. Actually, the states of neural concepts produced by the sub-network resulting from the unsupervised learning step (see Equation \eqref{eq:weight}) can be seen as approximations of patterns encoded by the features. These hidden patterns emerge from complex correlations and associations in the training data, which can be used to classify the instances.

Equation \eqref{eq:feature} shows how to calculate the relevance score of a feature from the inner and outer weights,

\begin{equation}
\label{eq:feature}
\Omega(f_i)= \sum_{j=1}^{M} \left|w_{ij}\right| + \sum_{j=1}^{N}\sum_{t=0}^{T} \left|r_{ij}^{(t)}\right|.
\end{equation}

The intuition of the relevance score is that important features will be represented by neural concepts having outgoing weights with large absolute values. In this measure, we consider the outgoing weights obtained during both learning phases (the unsupervised one that computes the inner weights and the supervised one that computes the weights connecting the features with the decisions). Notice that we did not include the bias weight matrices $B$ and $Q$ in this formula since they do not map to any features and would only capture noise. Likewise, we excluded the neurons' activation values from this calculation moved by the following assumption. If a neural concept is relevant but takes low activation values, then the pseudoinverse learning rule will compute weights with large absolute values. This is allowed in LTCN-based models since weights are neither constrained nor have any causal connotation. Actually, each weight must be analyzed by following the same statistical assumptions when interpreting coefficients in logistic regression models.  

\section{Numerical simulations}
\label{sec:simulations}

The following section is devoted to the numerical simulations and the ensuing discussion. Firstly, we present the research hypotheses and describe the datasets used for simulation purposes. Secondly, we study the effect of the nonlinear parameter on the classifier's performance. Thirdly, we compare our recurrence-aware classifier against state-of-the-art methods used to cope with structured classification problems. Finally, we illustrate how the feature relevance measure works in a case study. 

\subsection{Methodology and datasets}
\label{sec:simulations:datasets}

The experimentation methodology relies on four research hypotheses. Firstly, we claim that the performance of our recurrence-aware LTCN-based classifier is not affected when the network converges to the unique fixed-point attractor. Secondly, we claim that some problems can benefit from adding some linearity degree to the reasoning rule. Thirdly, we claim that the proposed neural classifier performs comparably to state-of-the-art black boxes. Finally, we claim that our LTCN-based classifier allows quantifying the relevance of each feature without the need for post-hoc procedures.

To investigate our research hypotheses, we adopt 30 pattern classification datasets taken from the study in \cite{Napolesb} and a case study concerning cybersecurity (to be presented in the last subsection). For the sake of convenience, we retained the datasets with numerical features and without missing values. Besides, all features have been normalized using the min-max scaling method. Table \ref{table:datasets} provides relevant information about these datasets, such as the number of instances, the number of features, the number of classes, and the imbalance ratio.   

\begin{table}
	\caption{Datasets used for simulation purposes.}
	\vspace{-2mm}
	\label{table:datasets}  
	\begin{center}
	\resizebox{\columnwidth}{!}{
	\begin{tabular}{|l|c|c|c|c|c|c|}
		\hline
		ID & Name & Instances & Features & Classes & Imbalance  \\  
		\hline
            D1 & banana & 5,300  & 2 & 2 & no  \\
            D2 & bank & 4,520 & 16 & 2 & 7:1  \\
            D3 & cardiotocography-10 & 2,126 & 19 & 10 & 11:1\\
            D4 & cardiotocography-3 & 2,126 & 35 & 3 & 10:1\\
            D5 & mfeat-factors & 2,000 & 216 & 10 & no \\
            D6 & mfeat-fourier & 2,000 & 77 & 10 & no \\
            D7 & mfeat-karhunen & 2,000 & 64 & 10 & no \\
            D8 & mfeat-morphological & 2,000 & 6 & 10 &  no\\
            D9 & mfeat-pixel & 2,000 & 240 & 10 & no \\
            D10 & mfeat-zernike & 2,000 & 25 & 10 &  no\\
            D11 & musk2 & 6,598 & 164 & 2 & no \\
            D12 & optdigits & 5,620 & 64 & 10 &  no\\
            D13 & page-blocks & 5,473 & 10 & 5 & 175:1\\
            D14 & pendigits & 10,992 & 13 & 10 &  no\\
            D15 & plant-margin & 1,600 & 64 & 100 & no \\
            D16 & plant-shape & 1,600 & 64 & 100 &  no\\
            D17 & plant-texture & 1,600 & 64 & 100 & no \\
            D18 & segment & 2,301 & 19 & 7 &  no\\
            D19 & spambase & 846 & 18 & 2 & no \\
            D20 & vehicle & 846 & 18 & 4 &  no\\
            D21 & vehicle0 & 846 & 18 & 2 &  no\\
            D22 & vehicle1 & 846 & 18 & 2 &  no\\
            D23 & vehicle2 & 846 & 18 & 2 &  no\\
            D24 & vehicle3 & 846 & 18 & 2 & no  \\
            D25 & waveform & 5,000 & 40 & 3 & no\\
            D26 & winequality-red & 1,599 & 11 & 6 & 68:1 \\
            D27 & winequality-white & 4,898 & 11 & 7 & 440:1 \\
            D28 & yeast & 1,484 & 8 & 10 & 93:1\\
            D29 & yeast1 & 1,484 & 8 & 2 & no \\
            D30 & yeast3 & 1,484 & 8 & 2 & 8:1 \\
    \hline
    \end{tabular}}
	\end{center}
\end{table}

In our experiments, we use the Cohen's kappa coefficient \cite{kappa1960} for measuring the classifiers' performance. This measure estimates the inter-rater agreement for categorical items and ranges in $[-1,1]$, where $-1$ indicates no agreement between the prediction and the ground-truth values, $0$ means no learning (i.e., random prediction), and $1$ means total agreement or perfect performance. While accuracy is considered mainstream, the kappa coefficient is a more robust measure since it considers the agreement occurring by chance, which is relevant for datasets with class imbalance \cite{japkowicz2011evaluating,ben2008comparison}. However, we will also report the accuracy values for the sake of completeness.

\subsection{Exploring the quasi-nonlinear model}
\label{sec:simulations:model}

This subsection studies the quasi-nonlinear reasoning model and illustrates the issues caused by the unique fixed-point attractor. For the simulations, we perform 5-fold cross-validation without any hyper-parameter tuning as we want to study the network's performance when varying nonlinearity and the number of iterations.

Figure \ref{fig:models} shows the kappa score obtained for each dataset when using the classic decision model (depicted in Figure \ref{fig:model1}) and our recurrence-aware decision model (depicted in Figure \ref{fig:model2}) for different $\phi$ values. In this experiment, the maximal number of iterations is set to 20 while neurons use a sigmoid transfer function.

\begin{figure}[!ht]
    \centering
    \includegraphics[width=0.5\textwidth]{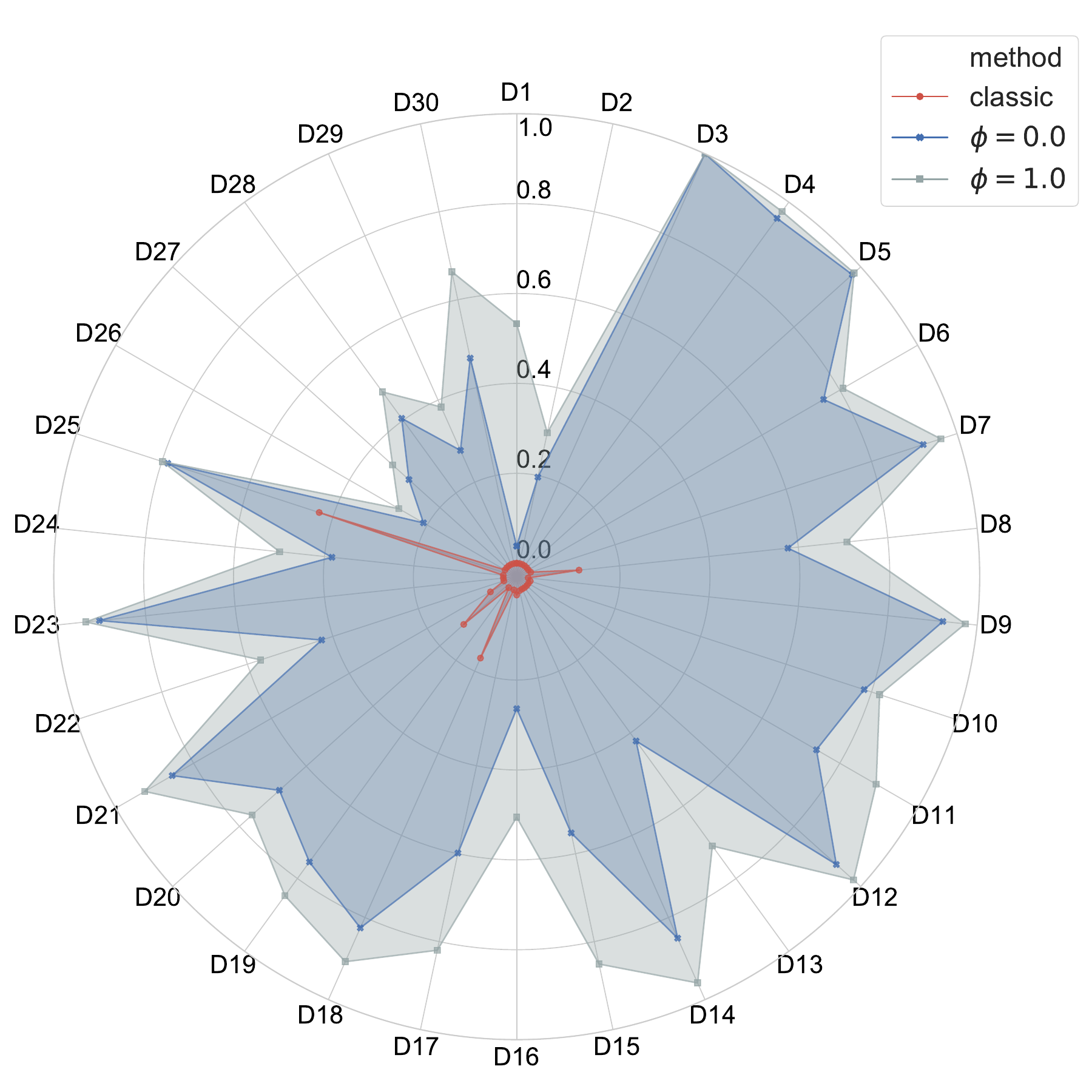}
    \captionsetup{justification=justified}
    \caption{Kappa scores obtained by the decision models depicted in Figures \ref{fig:model1} and \ref{fig:model2}. This simulation shows that our recurrence-aware model is not affected by the unique fixed point attractor regardless of the $\phi$ value. In contrast, the classic FCM-based classifier performs worse than a regression model ($\phi=0$).}
    \label{fig:models}
\end{figure}

Three conclusions can be drawn from this simulation. Firstly, the traditional decision model of FCM-based classifiers is worse than a regression model (that is to say, $\phi=0$) due to the convergence issues. It can be easily verified that the traditional FCM-based classifier converges to a unique fixed point, thus recognizing a single decision class (which causes the kappa value to be zero). Actually, the more iterations we perform, the higher the probability of observing such a behavior. Secondly, our recurrence-aware model is not affected by the unique fixed point regardless of the $\phi$ value. Thirdly, the models with larger $\phi$ values have better discriminatory capabilities. While this is expected, we should study the cases in which $0<\phi<1$ before jumping to definite conclusions.

\begin{figure*}[!htpb]
	\begin{subfigure}{0.33\textwidth}
		\includegraphics[height=4.8cm]{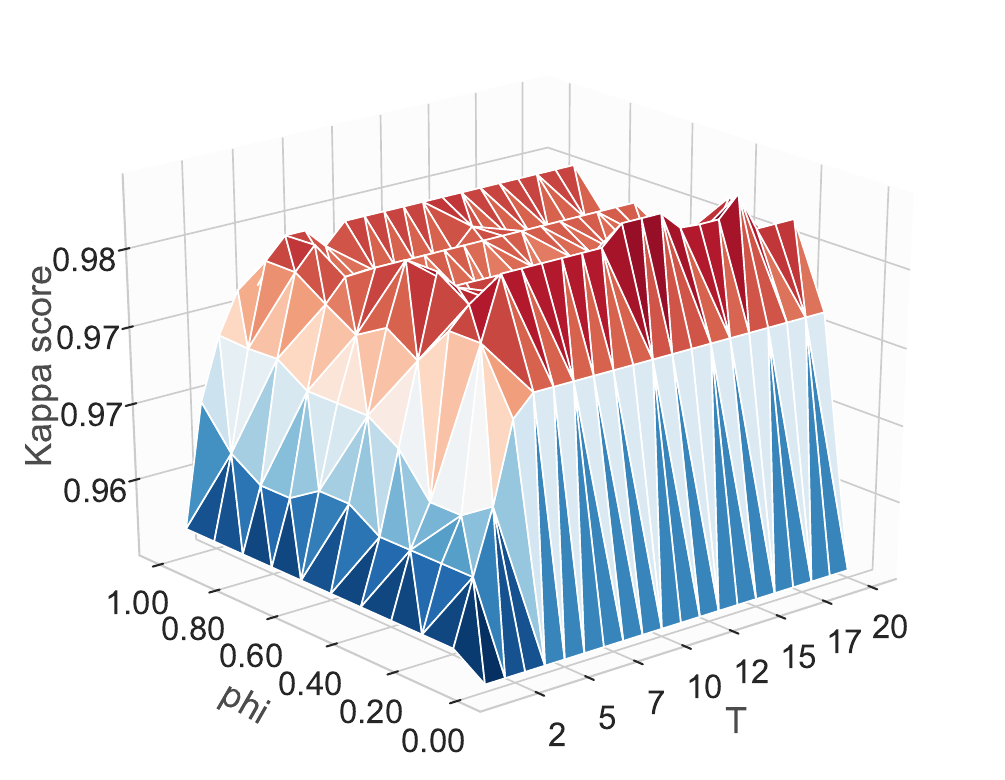}
		\caption{D4}
	\end{subfigure}
	\begin{subfigure}{0.33\textwidth}
		\includegraphics[height=4.8cm]{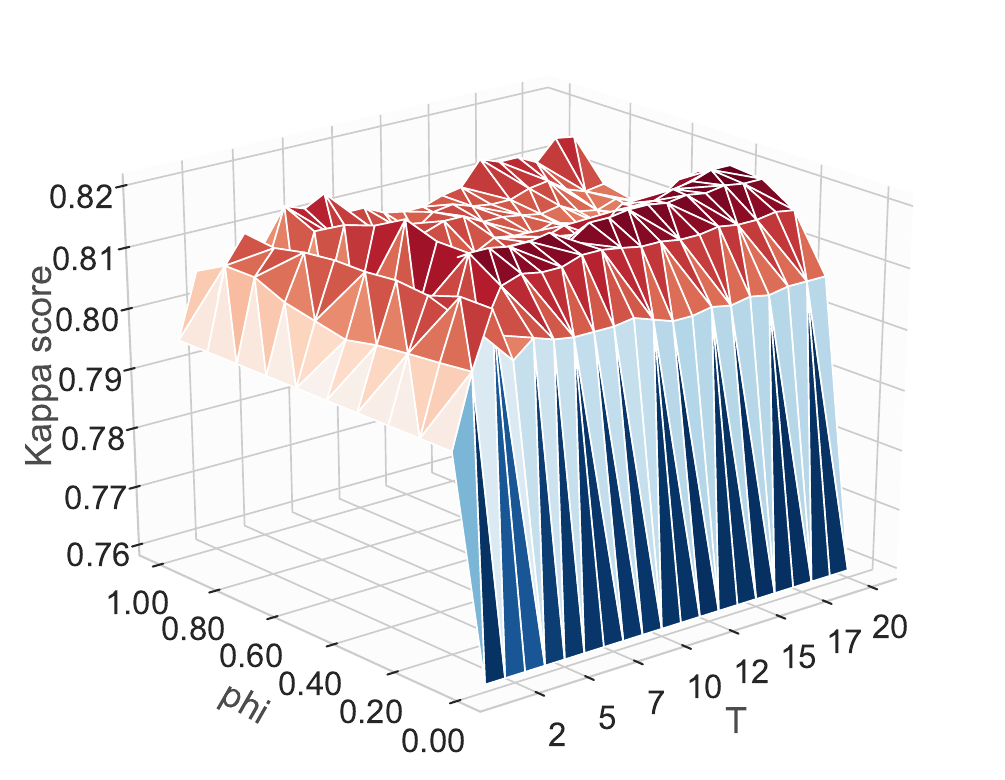}
		\caption{D6}
	\end{subfigure}
	\begin{subfigure}{0.33\textwidth}
		\includegraphics[height=4.8cm]{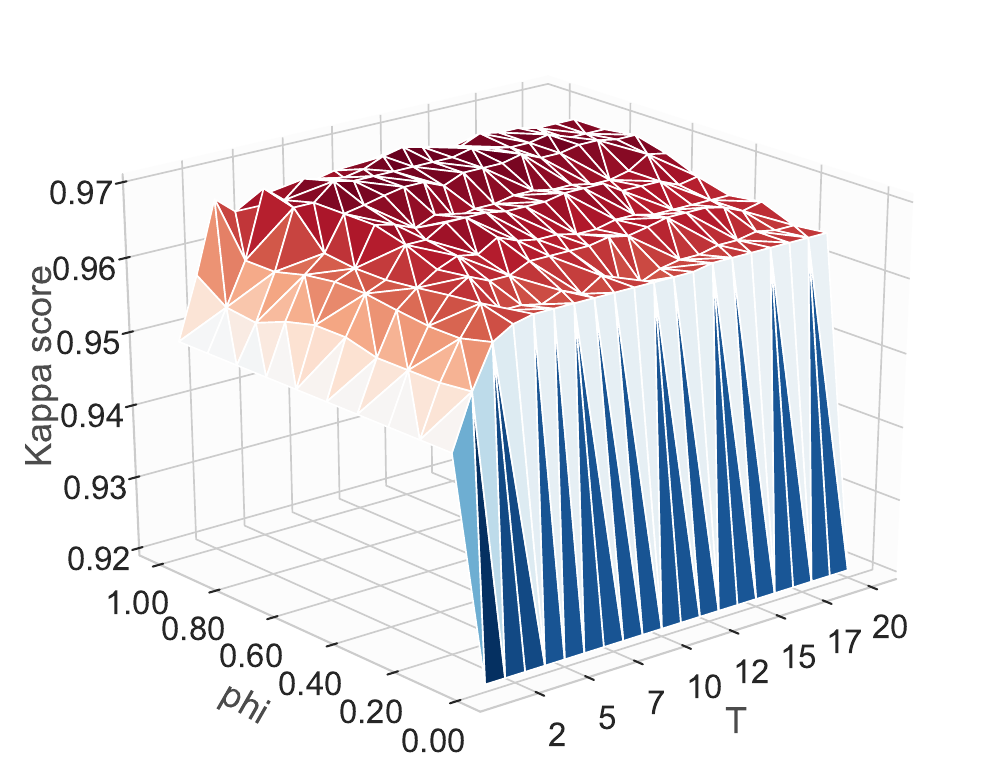}
		\caption{D7}
	\end{subfigure}
	\begin{subfigure}{0.33\textwidth}
		\includegraphics[height=4.8cm]{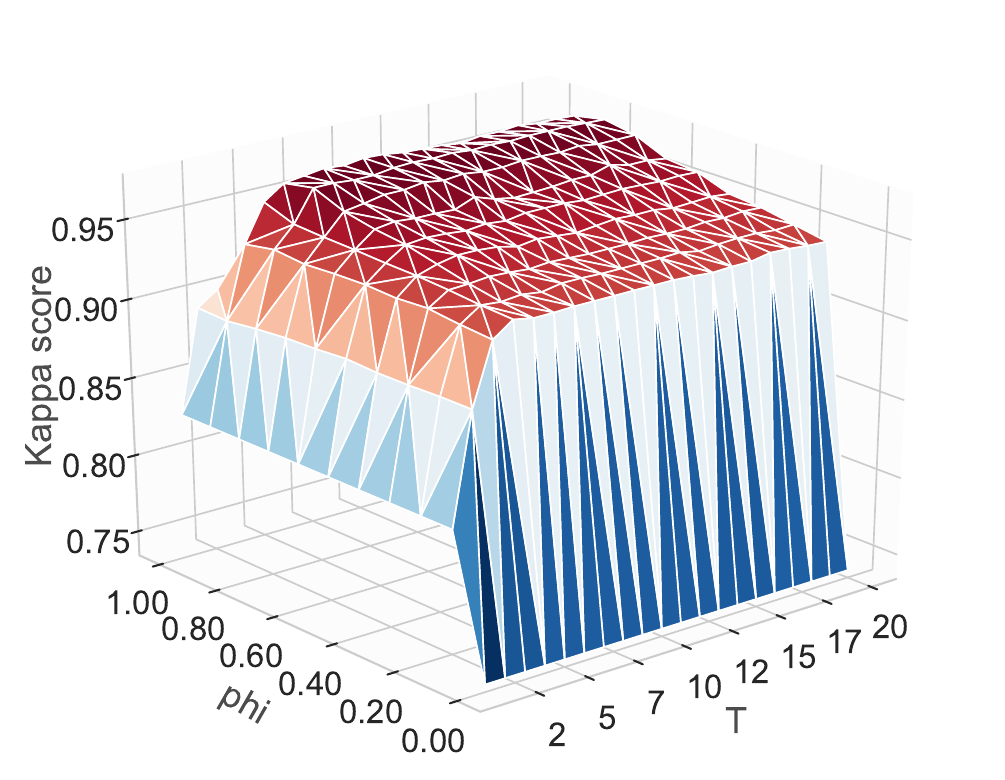}
		\caption{D11}
	\end{subfigure}
	\begin{subfigure}{0.33\textwidth}
		\includegraphics[height=4.8cm]{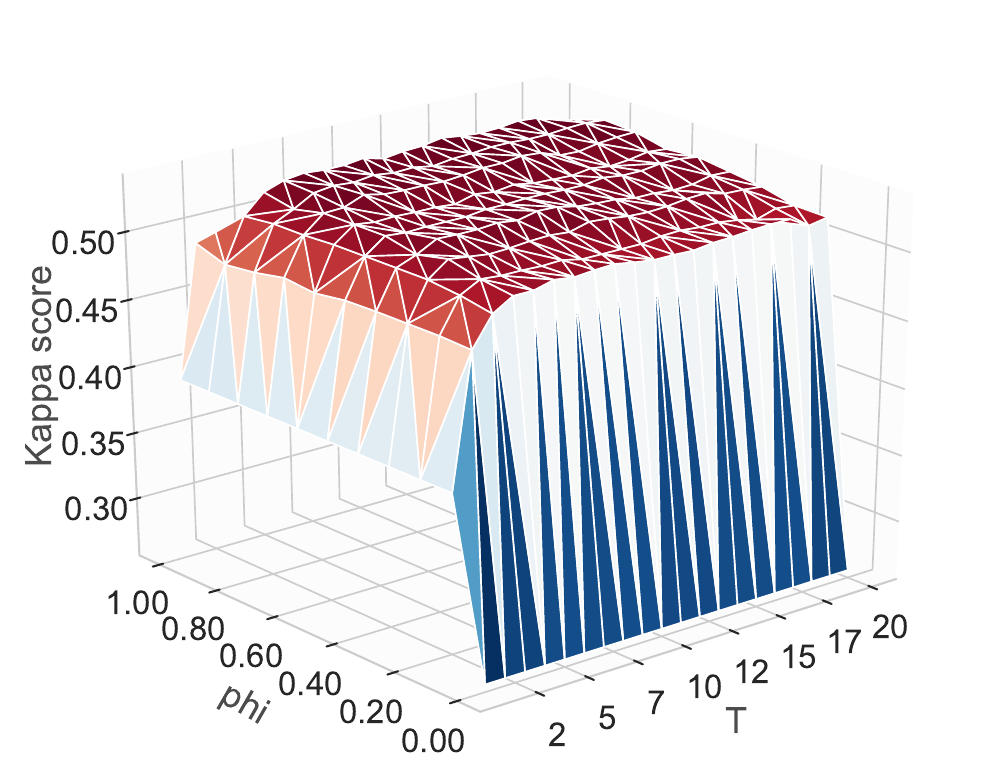}
		\caption{D16}
	\end{subfigure}
	\begin{subfigure}{0.33\textwidth}
		\includegraphics[height=4.8cm]{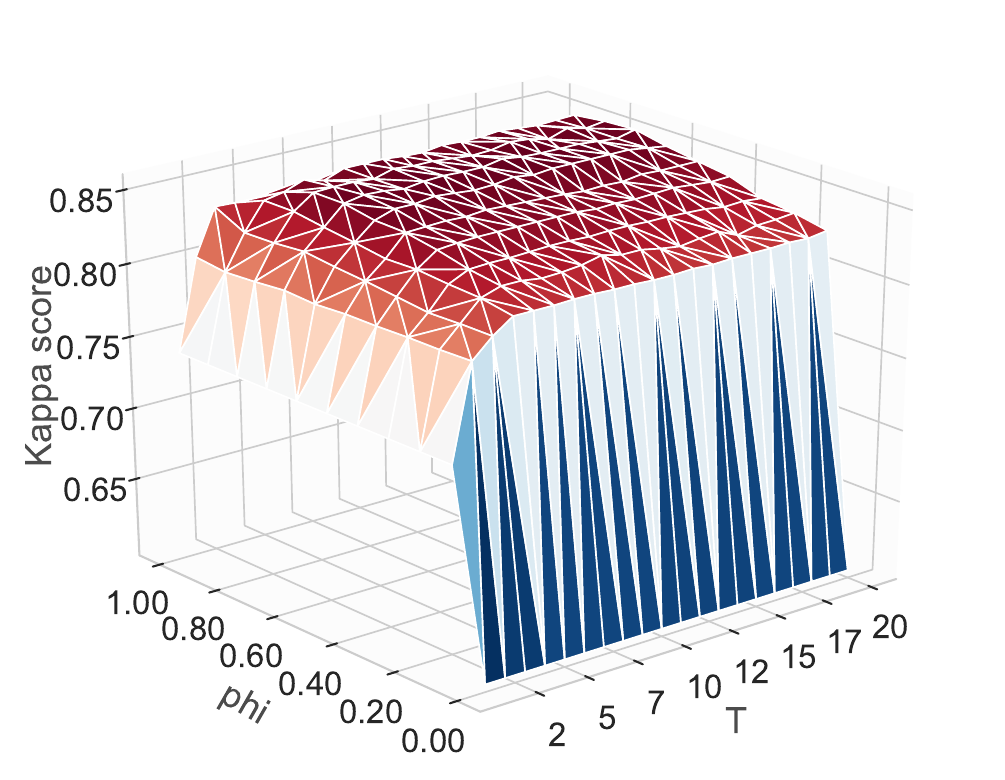}
		\caption{D17}
	\end{subfigure}
	\begin{subfigure}{0.33\textwidth}
		\includegraphics[height=4.8cm]{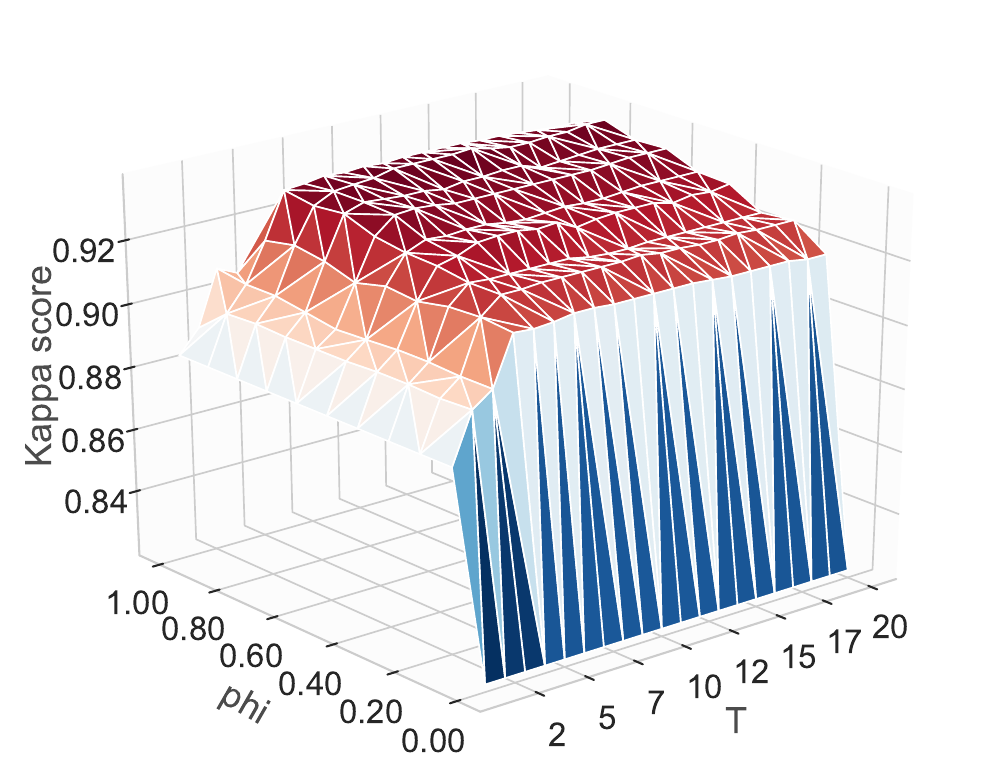}
		\caption{D18}
	\end{subfigure}
	\begin{subfigure}{0.33\textwidth}
		\includegraphics[height=4.8cm]{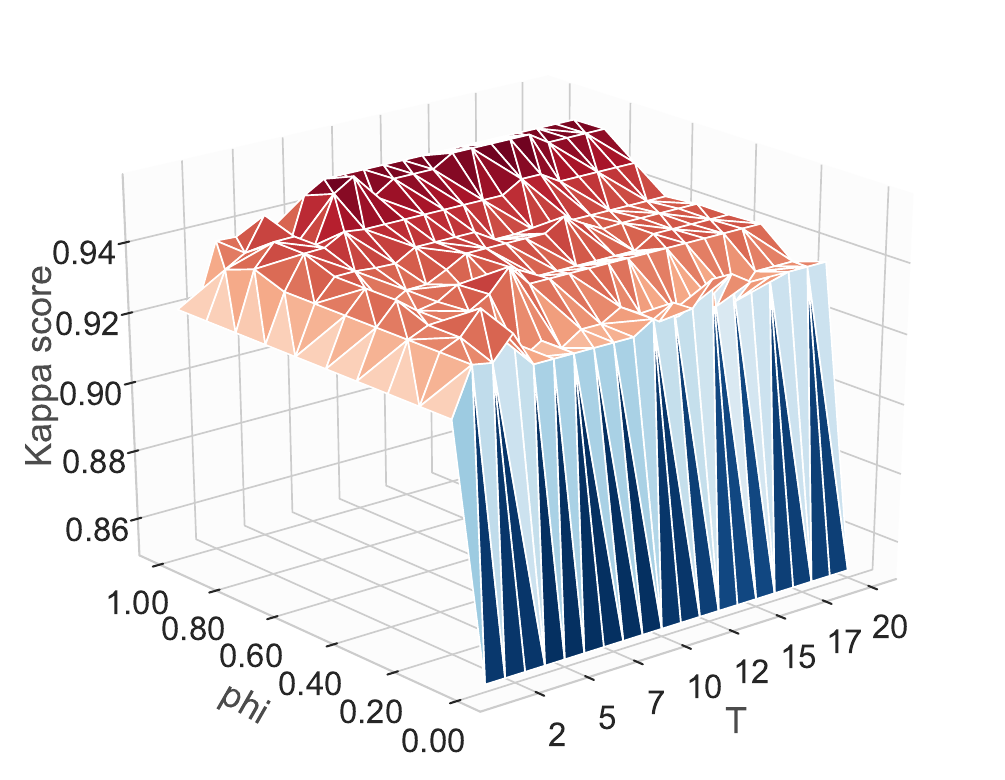}
		\caption{D21}
	\end{subfigure}
	\begin{subfigure}{0.33\textwidth}
		\includegraphics[height=4.8cm]{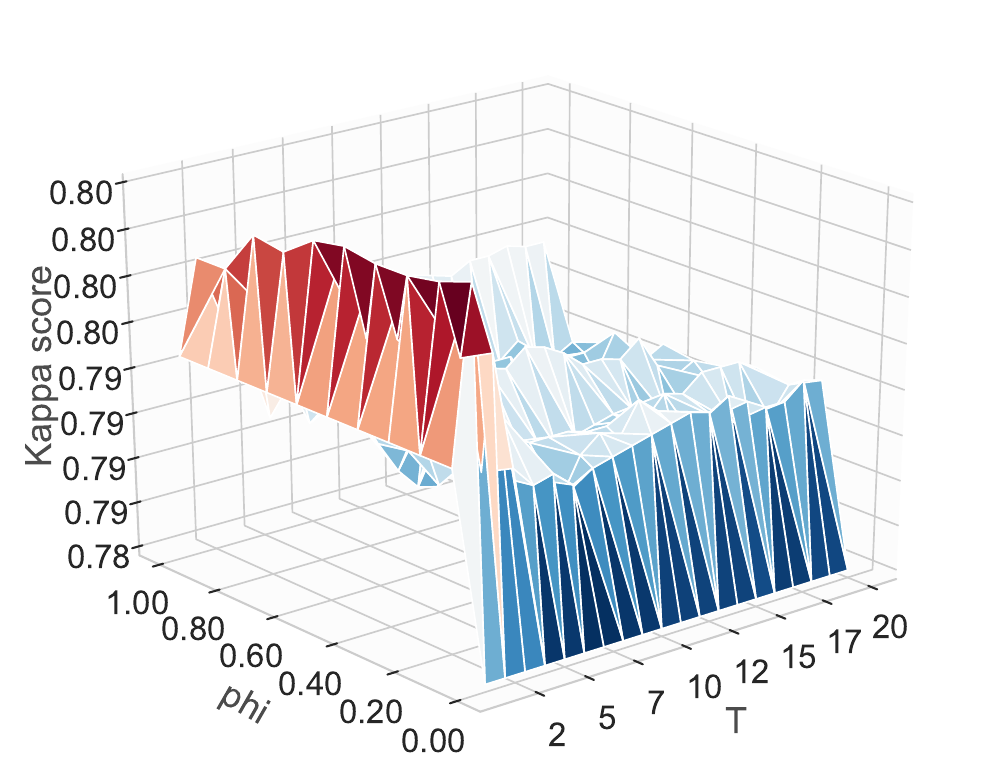}
		\caption{D25}
	\end{subfigure}
	\begin{subfigure}{0.33\textwidth}
		\includegraphics[height=4.8cm]{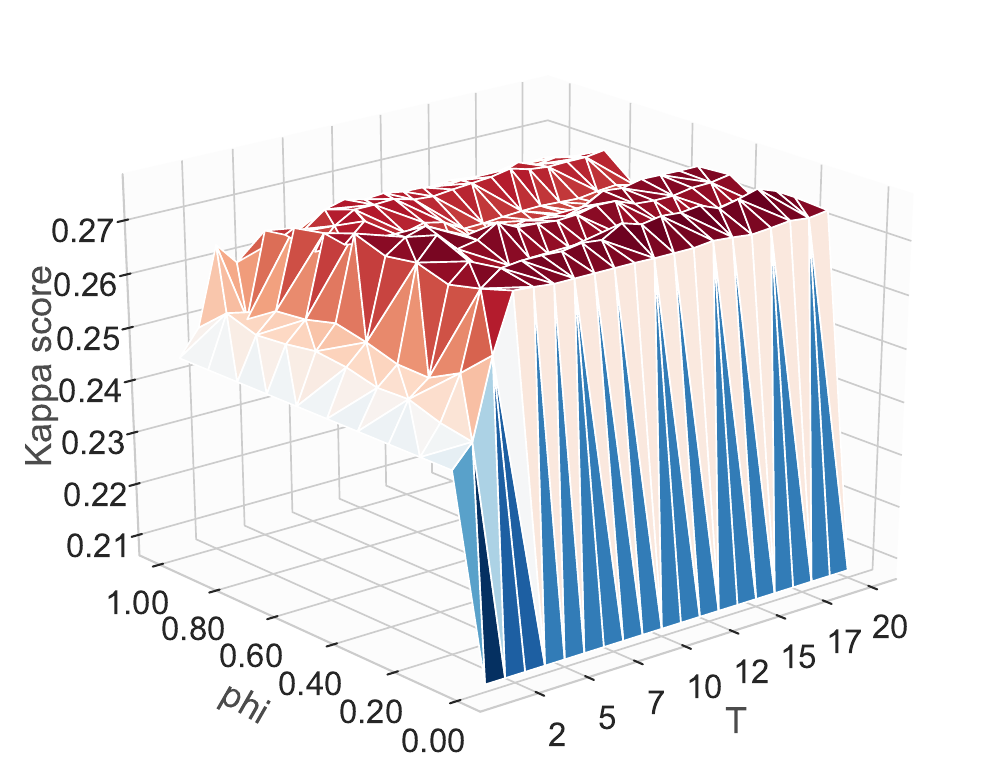}
		\caption{D27}
	\end{subfigure}
    \begin{subfigure}{0.33\textwidth}
		\includegraphics[height=4.8cm]{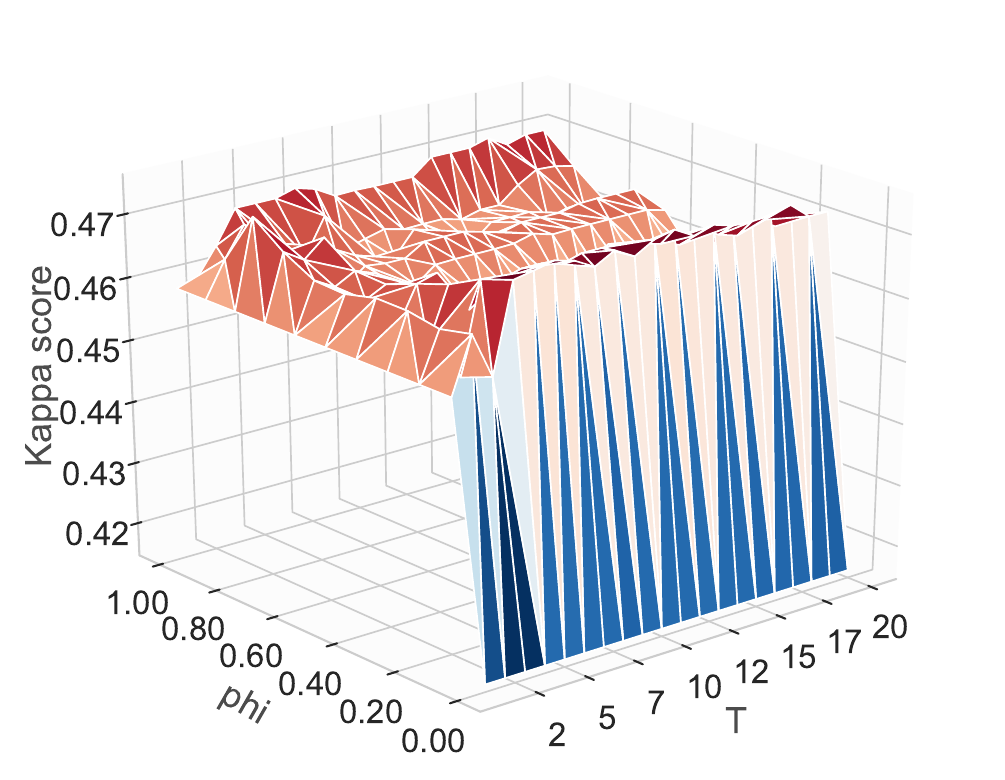}
		\caption{D28}
	\end{subfigure}
	\begin{subfigure}{0.33\textwidth}
		\includegraphics[height=4.8cm]{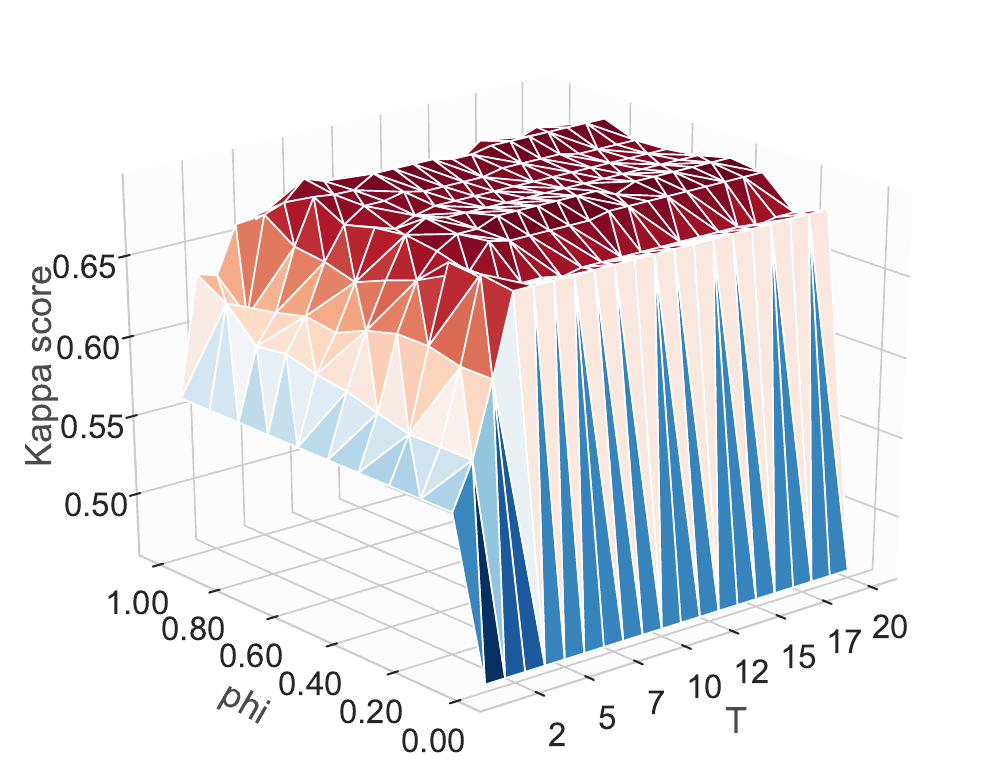}
		\caption{D30}
	\end{subfigure}

\captionsetup{justification=justified}
\caption{Kappa values obtained for selected datasets when varying $\phi$ and the number of iterations in our model. It can be noticed that $\phi=1$ does not necessarily yield the largest kappa score. Moreover, we can conclude that, for these problems, the performance does not increase much after performing five iterations. This happens because the network converges to a fixed point attractor, so more iterations do not add any new information to the augmented hidden state $H^{(t)}$ depicted in Equation \eqref{eq:decision}. In practice, we just stop the recurrent reasoning process when we notice the network has converged to a fixed-point attractor. In that way, we avoid adding the same columns to the hidden state $H^{(t)}$, which might cause issues when computing the Moore-Penrose pseudoinverse.}
\label{fig:parameters}
\end{figure*}

Figure \ref{fig:parameters} shows the kappa values obtained for selected datasets when varying the $\phi$ value and the number of iterations in our model. These figures support our second hypothesis: some problems can benefit from adding some linearity degree to the reasoning rule.

In addition, we perform an ablation study to determine the contribution of both learning processes: the unsupervised learning of the problem's dynamics and the supervised learning of the relations between the features and the decision classes. More specifically, we explored two scenarios: learning the unsupervised part while replacing the supervised learned weights with a random matrix, and learning the supervised part while replacing the unsupervised learned weights with a random matrix.

The average performance values for the first scenario are 0.8466 in terms of accuracy and 0.7305 in terms of kappa score. For the second scenario, the accuracy and kappa scores are even lower, equal to 0.2307 and -0.015, respectively. When comparing these results with LTCN including both learning processes (see table \ref{table:results_all}), the study shows that both parts of the learning process contribute to obtain the performance of our model. Learning the supervised weights has clearly more influence in the overall performance, but this is expected since the second part of the network is trained against the ground truth. However, the unsupervised learning part is fundamental for the intrinsic interpretability of the model.

\subsection{Exploring the model's predictive power}
\label{sec:simulations:predictive}

In this subsection, we contrast the performance of our model against state-of-the-art classifiers (both white and black boxes). The selected algorithms are Support Vector Machine (SVM), Logistic Regression (LR), Decision Tree (DT), Random Forest (RF) and Multilayer Perceptron (MLP) from the Scikit-learn library, Repeated Incremental Pruning to Produce Error Reduction (RIPPER) implemented in Weka, Self Organized Maps (SOM) \cite{Akinduko2016,Kohonen2003self}, Fuzzy-Rough Cognitive Networks (FRCN) \cite{Concepcion2020} and Efficient Gradient Boosting Tree (LightGBM) \cite{Ke2017}. These algorithms are able to cope with both binary and multi-class problems. In addition, we include Generalized Additive Model (GAM) \cite{serven2018} and Hybrid Rule Set (HyRS) \cite{Wang2021}, which can only cope with binary problems.


In our numerical experiments, we perform 5-fold nested cross-validation (i.e., with hyper-parameter tuning using the grid search method). Table \ref{table:hp_optimized} shows the hyper-parameters to be optimized and their values.

\begin{table}[!ht]
\caption{Hyper-parameters optimized}
	\label{table:hp_optimized}
	\resizebox{\columnwidth}{!}{
    \begin{tabular}{|l|l|}
    \hline
    Method                    & Hyper-parameters                                    \\ \hline
    \multirow{2}{*}{SVM}      & kernel = linear, poly, rbf, sigmoid;                \\
                              & C = 0.01 to 100;  gamma = scale, auto               \\ \hline
    \multirow{2}{*}{RF}       & number of estimators = 10 to 100;    \\
                              & criterion = gini, entropy; max\_depth = 2 to 10, \\
                              & sqrt(n\_inst)      \\ \hline
    \multirow{2}{*}{DT}       & criterion = gini, entropy; splitter = best, random; \\
                              & max\_depth = sqrt, log2, auto                       \\ \hline
    \multirow{3}{*}{MLP}      & transfer function = identity, logistic, tanh, relu; \\
                              & solver = lbfgs, sgd, adam; alpha = 0.01 to 0.5;     \\
                              & learning rate = constant, invscaling, adaptive;      \\ 
                              & hidden\_neurons = int((n\_att + n\_out) / 2) \\
                              \hline
    \multirow{2}{*}{RIPPER}   & folds = 2 to 10; optimizations = 2 to 10;           \\
                              & minimum total weight = 1.0 to 10.0                  \\ \hline
    \multirow{3}{*}{SOM}      & number of rows and columns = 10 to 100;             \\
                              & learning rate start = 0.5 to 0.9;                   \\
                              & number of iterations = 100 to 10,000                \\ \hline
    \multirow{3}{*}{FRCN}     & Implicator = Łukasiewicz, Gödel;                    \\
                              & T-norm = Łukasiewicz, Product;                     \\
                              & distance = HEOM, HMOM                               \\ \hline
    \multirow{2}{*}{GAM}      & number of splines = 5 to 10;                        \\
                              & fit splines = True or False;                        \\ \hline
    \multirow{2}{*}{LightGBM} & validation fraction = 0.1, 0.2;                     \\
                              & minimum samples per leaf = 10, 20                   \\ \hline
    \multirow{2}{*}{HyRS}     & max\_length = 3, 5; supp = 5, 10;                  \\
                              & number of rules = 1000, 5000                        \\ \hline
    LTCN                      & transfer function = sigmoid, tanh; phi = 0.0 to 1.0 \\ \hline
    \end{tabular}}
\end{table}

Table \ref{table:results_all} shows the average accuracy and kappa values of each classifier on the 30 datasets used in our experiments. The last column portrays the average training time (in seconds) of each model using an optimist approach based on the whole dataset since we just wanted to measure the training time. Table \ref{table:results_binary} reports the same statistics for binary classifiers. The simulations were run on a laptop with the following features: Core i9 7th generation, 10 physical processors, and 16GB RAM.

\begin{table}[!ht]
    \centering
    \caption{Average accuracy, kappa values and training time reported by each classifier after performing hyper-parameter tuning on all datasets.}
    \label{table:results_all}
    \begin{tabular}{|l|c|c|c|}
    \hline
        Method      & Accuracy  & Kappa    & Training Time      \\ \hline
        SVM         & 0.8733         & 0.7671        & 0.21   \\ \hline
        LR          & 0.8139         & 0.6256        & 0.09   \\ \hline
        DT          & 0.7686         & 0.6404        & 0.01   \\ \hline
        RF          & 0.8625         & 0.7522        & 0.53   \\ \hline
        MLP         & 0.8503         & 0.7278        & 0.19   \\ \hline
        RIPPER      & 0.9369         & 0.6669        & 1.95   \\ \hline
        SOM         & 0.5362         & 0.3376        & 2.98   \\ \hline
        FRCN        & 0.8500	     & 0.7395        & 65.59    \\ \hline
        LightGBM    & 0.8603	     & 0.7506        & 3.07  \\ \hline
        LTCN        & 0.8651         & 0.7613        & 0.14   \\ \hline
    \end{tabular}
\end{table}

\begin{table}[!ht]
    \centering
    \caption{Average accuracy, kappa values and training time reported by each classifier after performing hyper-parameter tuning on binary datasets.}
    \label{table:results_binary}
\begin{tabular}{|l|c|c|c|}
    \hline
        Method & Accuracy  & Kappa    & Training Time      \\ \hline
        GAM    & 0.8816         & 0.6448        & 10.94     \\ \hline
        HyRS   & 0.8770	        & 0.6117        & 18.87  \\ \hline
        LTCN   & 0.9163         & 0.7189        & 0.13   \\ \hline
    \end{tabular}
\end{table}

The simulation results show that LTCN, SVM, RF and LightGBM are the best-performing algorithms in terms of kappa scores, closely followed by MLP and FRCN, whereas SOM reports the worst results in our study. This comes with no surprise since these classifiers often report high prediction rates in tabular pattern classification problems. Simultaneously, it can be concluded that our algorithm's predictive power is far superior to the white boxes included in the study except for RIPPER. This method achieves higher accuracy, but its kappa score is lower than the kappa of our proposal, which means that it is less robust against datasets with high class imbalance.


It can be argued whether RF can still be considered a black box since we can quantify features' relevance as a proxy for interpretability. Despite this fact, our model involves other features, such as the possibility of domain experts injecting prior knowledge into the network. In other words, the expert can modify the inner weight matrix to encode rules that have not yet been observed in the data (i.e., the expected revenue increases after adding a new product to the stock). Achieving such a degree of flexibility with RF is not trivial.


Additionally, the experiments show that our method is the fastest among the most accurate algorithms. In summary, we can conclude that the proposed recurrence-aware LTCN classifier is as accurate as the black boxes included in the study while also being fairly fast.

\subsection{Exploring the model's interpretability}
\label{sec:simulations:interpretability}

In this subsection, we explore the interpretability of our neural model using a case study. The ``Phishing case” \cite{CHIEW2019153} is a binary classification problem described by 48 features extracted from 5000 phishing webpages and 5000 legitimate webpages and a class variable. Some of the features are: NumDots, UrlLength, NumDash, NumDashInHostname, AtSymbol, TildeSymbol, NumUnderscore, NumQueryComponents, NumAmpersand, NumHash, NumNumericChars, IpAddress, DoubleSlashInPath, PopUpWindow, ImagesOnlyInForm, and UrlLengthRT. 



The LTCN classifier reported an accuracy of 97\% after performing 5-fold cross-validation. In this experiment, we set the nonlinearity parameter to 0.8 and the number of iterations to 50 such that we can see what happens when we use more iterations than needed.

\begin{figure}[!ht]
    \centering
    \includegraphics[width=0.5\textwidth]{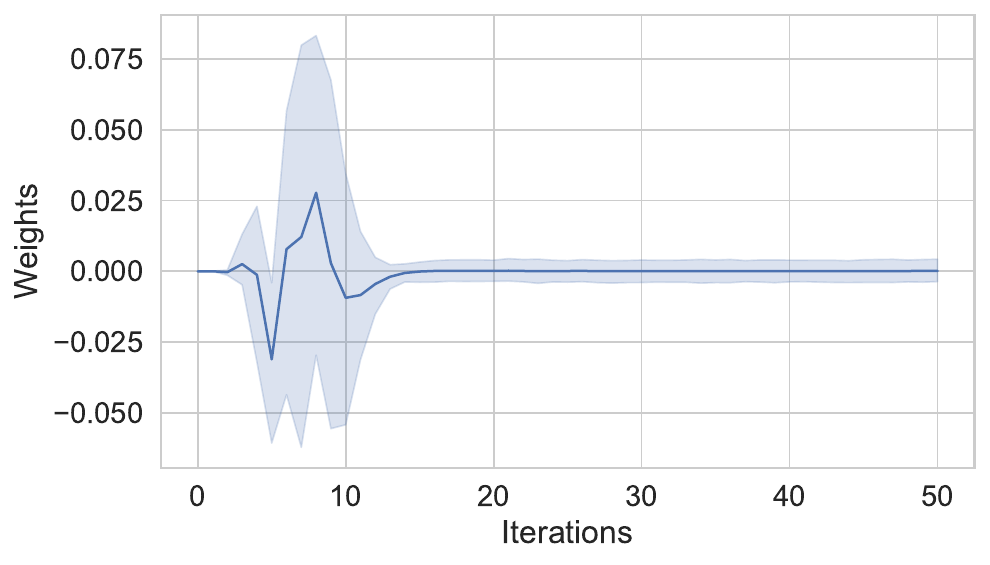}
    \captionsetup{justification=justified}
    \caption{Behavior of outer weights computed during the supervised learning step. The solid line represents the mean and the shadow indicates the 95\% confidence interval. It can be noticed that the weights connecting the inner neurons with the decision neurons do not change after some iterations.}
    \label{fig:phishing-weights}
\end{figure}

Figure \ref{fig:phishing-weights} shows the behavior of outer weights connecting the temporal states with the decision neurons. In some iterations, the supervised learning algorithm estimates the same weights repeatedly. This happens because the LTCN converges to a fixed point where the network's state does not change as the iterations continue.

\begin{figure}[!ht]
    \centering
    \includegraphics[width=0.5\textwidth]{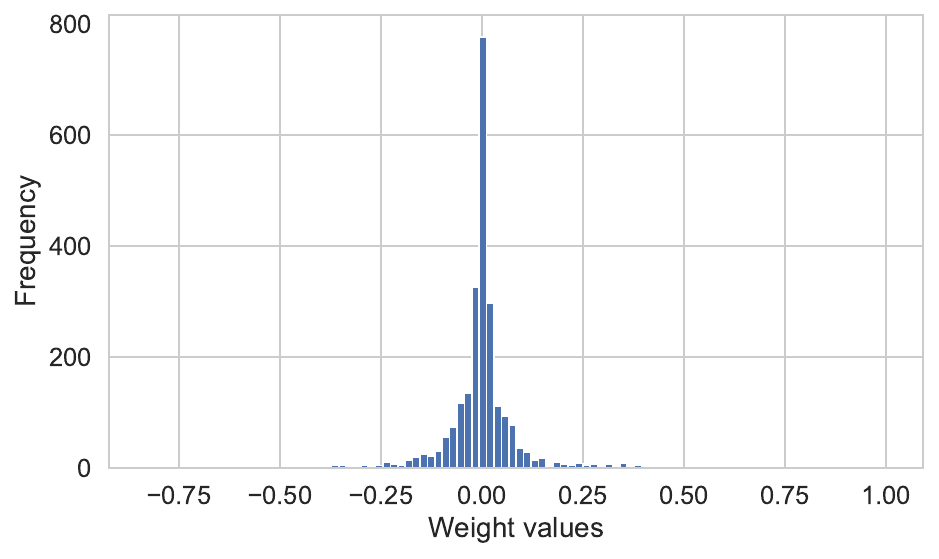}
    \captionsetup{justification=justified}
    \caption{Distribution of outer weights. The weights follow a zero-mean Gaussian distribution, meaning that the learning method is able to produce sparse weight matrices.}
    \label{fig:phishing-histogram}
\end{figure}

Figure \ref{fig:phishing-histogram} displays the histogram for the normalized outer weights. It can be noted that weights follow a zero-mean normal distribution, which suggests that the learning algorithm based on the Moore-Penrose pseudoinverse learns sparse weights. This desired behavior is unexpected since we did not consider any regularization component when designing the learning algorithm.  

\begin{figure}[!ht]
    \centering
    \includegraphics[width=0.5\textwidth]{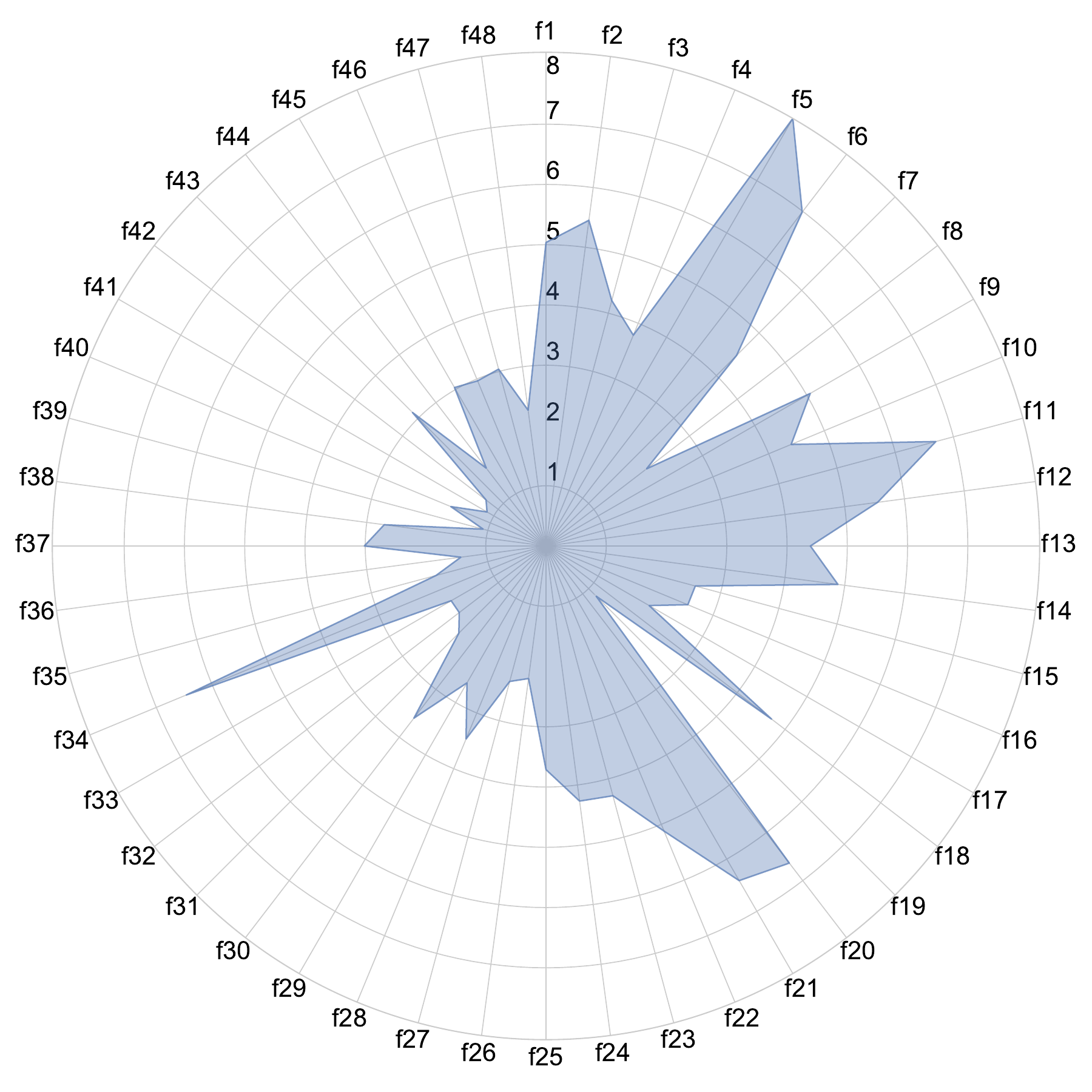}
    \captionsetup{justification=justified}
    \caption{Importance scores of problem features. The fifth feature (NumDash) seems to be of utmost importance for the decision model of our FCM-based classifier.}
    \label{fig:feature-importance}
\end{figure}

Next, we compute the relevance score attached to each feature using Equation \eqref{eq:feature}. As mentioned, the intuition of this measure is that important features (represented with meaningful neural concepts in the network) will be connected through outgoing weights with large absolute values. Figure \ref{fig:feature-importance} portrays the relevance scores such that $f_i$ denotes the $i$-th problem feature. The largest relevance score corresponds to $f_5$ (NumDash). Other important features are: $f_6$ (NumDashInHostname), $f_{11}$ (NumQueryComponents), $f_{20}$ (HttpsInHostname), $f_{21}$ (HostnameLength), and $f_{34}$ (PctNullSelfRedirectHyperlinks).

Validating these scores is not an easy task: different models might focus on different features. That is why having a comprehensible measure is important to gain trustworthiness in the results \cite{Juang2016}. Despite this fact, we insisted on comparing the relevance scores with the coefficients of a logistic model after reducing the number of features (for the sake of simplicity). Aiming at selecting the features, we used the \texttt{CfsSubsetEval} method \cite{Hall1998}, which evaluates the worth of a subset of features by considering the individual predictive ability of each feature along with the degree of redundancy between them.

The selected features were $f_1$ (NumDots), $f_3$ (PathLevel), $f_5$ (NumDash), $f_{14}$ (NumChars), $f_{25}$ (NumSensitiveWords), $f_{27}$ (PctExtHyperlinks), $f_{30}$ (InsecureForms), $f_{34}$ (PctNullSelfRedirectHyperlinks), $f_{35}$ (FrequentDomainNameMismatch), $f_{38}$ (PopUpWindow), $f_{39}$ (SubmitInfoToEmail), $f_{47}$ (ExtMetaScriptLinkRT), and $f_{48}$ (PctExtNullSelfRedirectHyperlinksRT).

Equation \eqref{eq:logistic} depicts the logistic regression model obtained from these features,

\begin{multline}
\label{eq:logistic}
g(x) = -6.83f_1 - 7.62f_3 + \textbf{26.17}f_5 - 0.41f_{14} - 5.53f_{25}\\ + 8.98f_{27} - 3.14f_{30} + 4.51f_{34} - 4.11f_{35} + 4.5f_{38} + \\ 3.79f_{39} - 2.37f_{47} + 9.37f_{48} -3.0917.
\end{multline}

While the weights of selected features differ from those reported by our measure, the $f_5$ feature continues to have the largest weight in the logistic regression model. We also observed an interesting behavior when using all features to build a regularized regression model: the larger the penalization, the larger the weight of $f_5$. This further confirms that this feature is quite important for the decision process of these regression-like classifiers.

\section{Concluding remarks}
\label{sec:conclusions}

In this paper, we presented a recurrence-aware LTCN-based model for interpretable pattern classification. While many FCM-based classifiers have the limitation of converging to a unique fixed point, therefore being able to recognize only one decision class, our proposal evades this by establishing temporal connections between all earlier states and the decision neurons. In that way, our model focuses on the trajectory to the equilibrium point rather than the fixed point itself. In addition, we proposed a quasi-nonlinear reasoning rule to control nonlinearity. Another drawback found in the literature is that many learning algorithms for FCM-like models are meta-heuristic-based and suffer from being slow. This proposal dodges these issues by employing instead a two-step deterministic learning method. Last but not least, we propose a feature relevance score as a proxy for interpretability.

The numerical simulations using 30 structured pattern classification problems supported our research hypotheses. First, we confirmed that our recurrence-aware LTCN model is not affected by the unique fixed point attractor problem, in contrast to the FCM-based classifier, which performed worse than a logistic regression model. Second, we illustrated how some problems benefit from models having a certain linearity degree (i.e., using the neurons' initial states when computing the temporal states). Third, we showed that our model's discriminatory capability is comparable to state-of-the-art algorithms. Moreover, we noted that the learning method produced sparse weight representations, even when we did not consider any regularization strategy. Finally, it was found that our feature relevance measure aligns well with the interpretability derived from a logistic regression model. However, it is worth recalling that different models might focus on different features without reporting any contradiction.

Within the limitations of our proposal, we should mention that the whole approach might lead to memory issues when solving problems characterized by thousands of features, which typically need a larger number of iterations to produce accurate results. Our next research step will be devoted to sorting out this issue.

\section*{Acknowledgment}
Y. Salgueiro would like to acknowledge the support provided by the Program CONICYT FONDECYT de Postdoctorado through project 3200284. This research was partially supported by the super-computing infrastructure of the NLHPC (ECM-02). We thank the anonymous reviewers for their constructive criticism, valuable comments and suggestions.

\bibliographystyle{IEEEtran}



\end{document}